\documentclass[11pt,letterpaper]{mystyle}
\usepackage{fancyhdr}

\usepackage[numbers]{natbib}
\usepackage{adjustbox}
\usepackage{hyperref}    
\usepackage[edges]{forest}
\usepackage{subcaption}
\usepackage{soul}
\usepackage{multirow}

\usepackage{booktabs}       
\usepackage{colortbl}
\usepackage{enumitem}
\usepackage{amssymb}
\usepackage{wrapfig}
\usepackage{courier}
\usepackage{bxcoloremoji}
\usepackage{CJKutf8}

\usepackage[utf8]{inputenc} 
\usepackage[T1]{fontenc}    
\usepackage{hyperref}       
\usepackage{url}            
\usepackage{booktabs}       
\usepackage{amsfonts}       
\usepackage{nicefrac}       
\usepackage{microtype}      
\usepackage{xcolor}         

\usepackage{amsmath}
\usepackage{tabularx}
\usepackage{wrapfig}
\usepackage{enumitem}
\usepackage{forest}
\usepackage{xcolor}
\usepackage{textcomp}
\usepackage{amsmath}
\usetikzlibrary{arrows.meta,positioning}

\definecolor{memory-tree}{HTML}{3d92cf}
\definecolor{tooling-tree}{HTML}{42b98f}  
\definecolor{planning-tree}{HTML}{7051c8}
\definecolor{benchmark-tree}{HTML}{bb5e2b}

\definecolor{color1}{rgb}{0.1,0.7,0.8}
\definecolor{color2}{rgb}{0.9,0.1,0.1}
\definecolor{color3}{rgb}{0.7,0.3,0.7}
\definecolor{color4}{rgb}{0.3,0.3,0.7}
\definecolor{color5}{RGB}{8, 102, 3}
\definecolor{color6}{rgb}{0.53, 0.66, 0.42}

\definecolor{SoftBlue}{RGB}{88, 130, 255} 
\tcbset{
  infoboxstyle/.style={
    colframe=blue!75!black, colback=blue!10!white
  }
}
\usepackage{fontawesome5}
\usepackage{hyperref}

\newcommand{\ghlink}[1]{\faIcon{github}\,\href{#1}{GitHub}}
\newcommand{\weblink}[1]{\faIcon{globe}\,\href{#1}{Website}}
\newtcolorbox{infobox}[1]{infoboxstyle={#1}}

\newcommand{\github}{\raisebox{-1.5pt}{\includegraphics[height=1.05em]{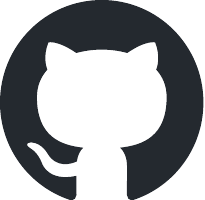}}}
\newcommand{\hflink}[1]{%
\includegraphics[height=1.em]{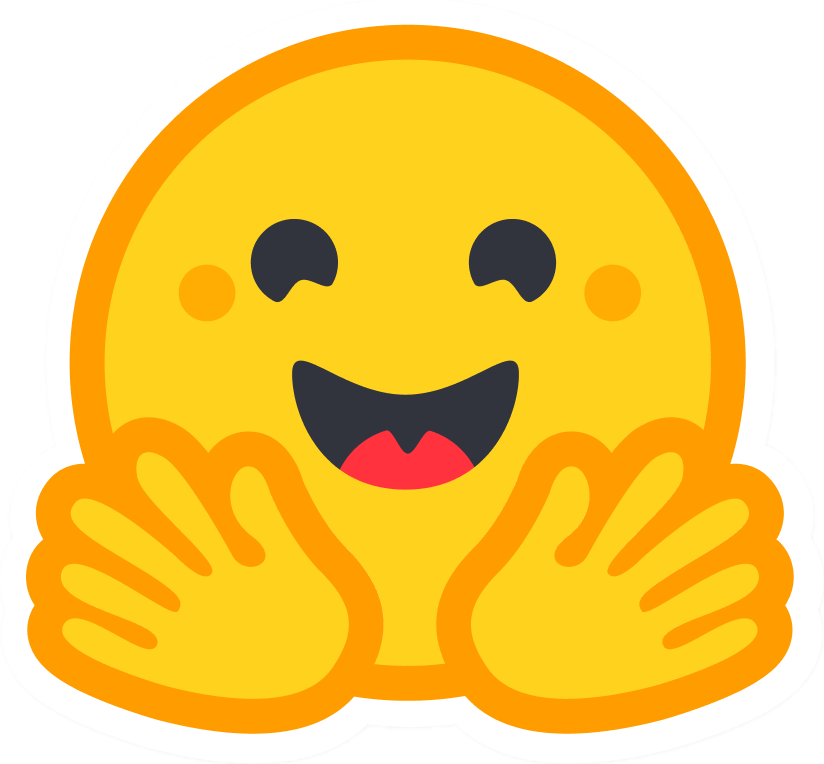}\,\href{#1}{HuggingFace}%
}

\newtcolorbox{summarybox}[2][]{
    colback = blue!5!white, 
    colframe = blue!50!white, 
    coltitle = black,
    fonttitle = \bfseries,
    colbacktitle = blue!5!white, 
    enhanced,
    attach boxed title to top left={xshift=2mm, yshift=-4mm},
    top=5mm,
    title=#2,
    #1
}

\definecolor{hidden-red}{RGB}{205, 44, 36}
\definecolor{hidden-blue}{RGB}{194,232,247}
\definecolor{hidden-orange}{RGB}{243,202,120}
\definecolor{hidden-green}{RGB}{34,139,34}
\definecolor{hidden-pink}{RGB}{255,245,247}
\definecolor{hidden-black}{RGB}{20,68,106}
\definecolor{purple}{RGB}{144,153,196}
\definecolor{yellow}{RGB}{255,228,123}
\definecolor{hidden-yellow}{RGB}{255,248,203}
\definecolor{tkcolor}{RGB}{224,223,255}
\definecolor{darkblue}{rgb}{0, 0.40, 0.75}

\hypersetup{colorlinks=true, citecolor=darkblue, linkcolor=darkblue, urlcolor=darkblue}
\tcbset{
  aibox/.style={
    width=\linewidth,
    top=8pt,
    bottom=4pt,
    colback=blue!6!white,
    colframe=black,
    colbacktitle=black,
    enhanced,
    center,
    attach boxed title to top left={yshift=-0.1in,xshift=0.15in},
    boxed title style={boxrule=0pt,colframe=white,},
  }
}

\newtcolorbox{AIbox}[2][]{aibox,title=#2,#1}

\tcbset{
  takeawaybox/.style={
    width=\linewidth,
    top=8pt,
    bottom=4pt,
    colback=hidden-yellow,
    colframe=black,
    colbacktitle=black,
    enhanced,
    center,
    attach boxed title to top left={yshift=-0.1in,xshift=0.15in},
    boxed title style={boxrule=0pt,colframe=white,},
  }
}

\newtcolorbox{TakeawayBox}[2][]{takeawaybox,title=#2,#1}

\title{Toward Efficient Agents: A Survey of Memory, Tool Use, and Planning}

\author{
    Xiaofang Yang$^{1,2,\text{\textdagger}}$ \quad Lijun Li$^{1,\text{\textdagger},{\coloremojicode{2709}}}$ \quad Heng Zhou$^{1,3,\text{\textdagger}}$ \quad Tong Zhu$^{1,\text{\textdagger}}$ \quad Xiaoye Qu$^{1}$ \quad Yuchen Fan$^{1,4}$ \quad Qianshan Wei$^5$ \quad Rui Ye$^4$ \quad Li Kang$^{1,4}$ \quad Yiran Qin$^6$ \quad Zhiqiang Kou$^7$ \quad Daizong Liu$^8$ \quad Qi Li$^5$ \quad Ning Ding$^9$ \quad Siheng Chen$^4$ \quad Jing Shao$^{1, {\coloremojicode{2709}}}$ \\
    \normalfont{
        $^1$ Shanghai Artificial Intelligence Laboratory,
        $^2$ Fudan University, \vspace{-5pt}\\
        $^3$ University of Science and Technology of China, 
        $^4$ Shanghai Jiaotong University,\vspace{-5pt}\\
        $^5$ Institute of Automation, Chinese Academy of Sciences, \vspace{-5pt}\\
        $^6$ The Chinese University of Hong Kong (Shenzhen), \vspace{-5pt}\\
        $^7$ Hong Kong Polytechnic University, $^8$ Wuhan University,
        $^9$ Tsinghua University
        \vspace{-5pt}\\
        
    }
}

\begin{document}

\begin{abstract}
  \vspace{5mm}
  \textbf{\large Abstract:}
  \vspace{2mm}

Recent years have witnessed increasing interest in extending large language models into agentic systems. While the effectiveness of agents has continued to improve, \textbf{efficiency}, which is crucial for real-world deployment, has often been overlooked. This paper therefore investigates efficiency from three core components of agents: \textbf{memory, tool use, and planning}, considering costs such as latency, tokens, steps, etc. 
Aimed at conducting comprehensive research addressing the efficiency of the agentic system itself, we review a broad range of recent approaches that differ in implementation yet frequently converge on \textbf{shared high-level principles} including but not limited to bounding context via compression and management, designing reinforcement learning rewards to minimize tool invocation, and employing controlled search mechanisms to enhance efficiency, which we discuss in detail. 
Rather than treating efficiency as a standalone objective, we adopt an effectiveness-first perspective that an agent is meaningfully efficient only when it preserves acceptable task quality. Accordingly, we characterize efficiency in two complementary ways: comparing effectiveness under a fixed cost budget, and comparing cost at a comparable level of effectiveness. 
From this perspective, we survey benchmarks along both dimensions. We first review effectiveness-oriented benchmarks across holistic and component-specific evaluations, and then examine how efficiency is measured in practice, summarizing evaluation protocols and commonly reported metrics from both benchmark papers and method-level studies.
Moreover, we discuss the key challenges and future directions, with the goal of providing promising insights.
  \vspace{5mm}

  $^{\text{\textdagger}}$ \textit{Main contributors}

  $^{\coloremojicode{2709}}$ \textit{Corresponding Author}

  \vspace{5mm}
  \vspace{5mm}

  \coloremojicode{1F5D3} \textbf{Date}: January 20th, 2026

  \coloremojicode{1F3E0} \textbf{Projects}: \href{https://efficient-agents.github.io/}{https://efficient-agents.github.io/}

  \github \textbf{Code Repository}: \href{https://github.com/yxf203/Awesome-Efficient-Agents}{https://github.com/yxf203/Awesome-Efficient-Agents}

  \coloremojicode{1F4E7} \textbf{Contact}: \href{mailto:yangxiaofang@pjlab.org.cn}{yangxiaofang@pjlab.org.cn}, \href{mailto:lilijun@pjlab.org.cn}{lilijun@pjlab.org.cn}, \href{mailto:hengzzzhou@gmail.com}{hengzzzhou@gmail.com}, \href{mailto:zhutong@pjlab.org.cn}{zhutong@pjlab.org.cn}

\end{abstract}
\maketitle

\vspace{3mm}
\thispagestyle{empty}
\newpage
\tableofcontents

\newpage
\section{Introduction}
\begin{figure}[ht]
    \centering
    \includegraphics[width=\textwidth]{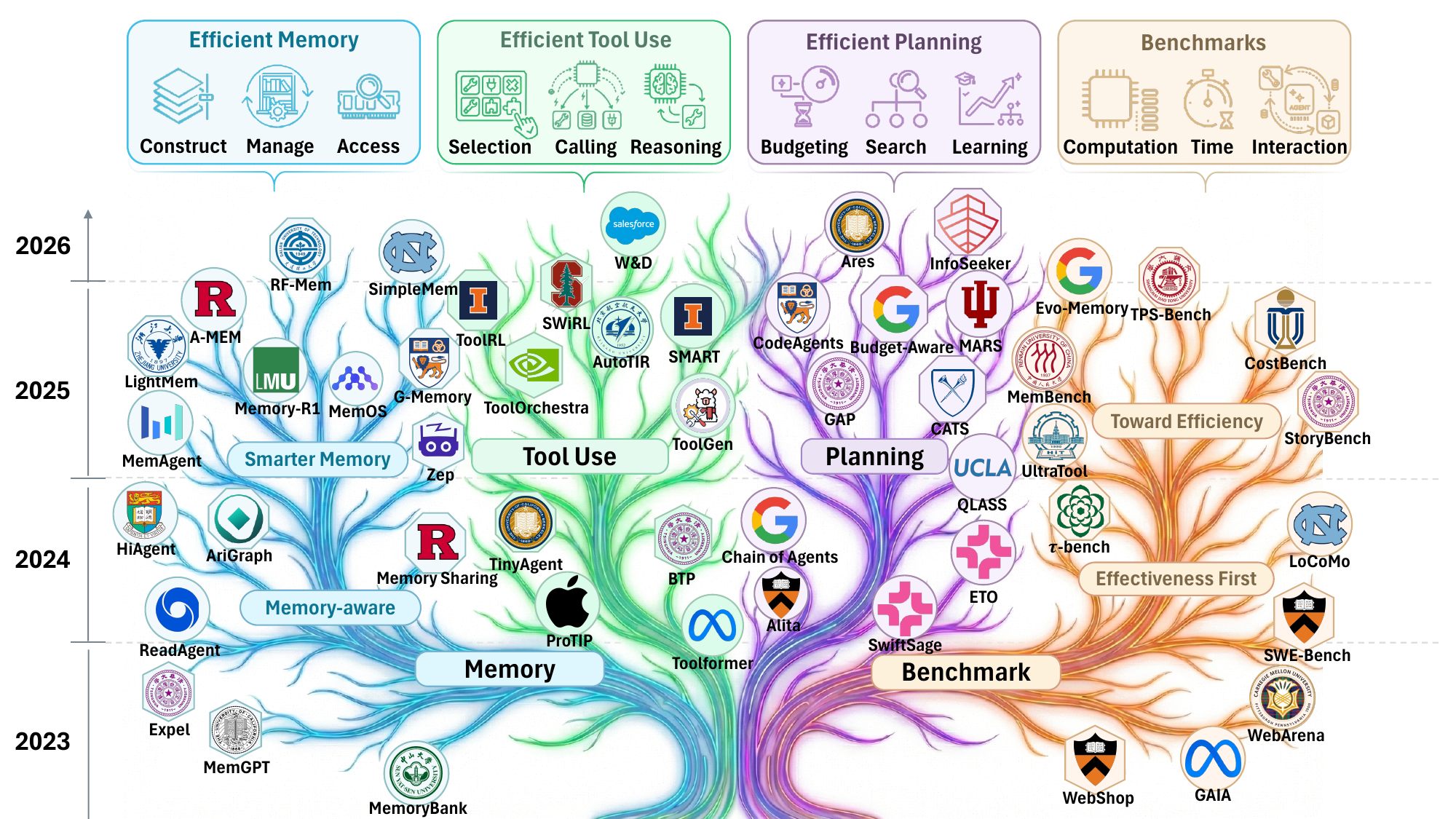}  
    \caption{The evolutionary trajectory of efficient agent research. The diagram is organized into four principal branches: \textcolor{memory-tree}{Memory}, \textcolor{tooling-tree}{Tool Use}, \textcolor{planning-tree}{Planning}, and \textcolor{benchmark-tree}{Benchmarks}. Key works and their institutional affiliations are mapped chronologically to illustrate the field's development and categorization from 2023 to 2026.}
    \label{fig:evolution}
\end{figure}
The landscape of Artificial Intelligence has undergone a paradigm shift, evolving from the era of Convolutional Neural Networks (CNNs) and Recurrent Neural Networks (RNNs) to the advent of Large Language Models (LLMs), and the emergence of LLM-based Agents currently~\cite{cnn,rnn,rnn1,gpt,agent1,agent2}. Unlike their predecessors, which primarily focused on perception or static text generation, agentic systems do not merely process information; they actively interact with external environments to execute complex, multi-step workflows across diverse domains, such as autonomous software engineering~\cite{code1,code2} and accelerated scientific discovery~\cite{scientist1,scientist2,scientist3}.

However, this shift toward autonomous action has introduced a critical bottleneck: \textbf{efficiency}. While the deployment of LLMs is already resource-intensive, this challenge is \textbf{significantly exacerbated} in agentic systems. Unlike a standard LLM that typically operates in a linear, single-turn query-response format, an agent consumes exponentially more resources due to its recursive nature. To automate intricate real-world tasks~\cite{cogagent,webvoyager,alphaevolve,code2}, agents must perform extensive memory management, iterative tool usage, and complex planning over multiple steps. This multi-step execution leads to prohibitive latency, context window saturation, and excessive token consumption, raising profound concerns regarding the long-term sustainability and equitable accessibility of these increasingly capable systems.

To understand the urgency of agent efficiency, one must examine the typical agentic workflow. Upon receiving a user instruction, an agent engages in a recursive loop that heavily uses the following key components: memory, planning, and tool use to observe output and provide the final solution.
\begin{equation*}
\begin{aligned}
\mathrm{Input}\rightarrow \Bigl[\,
\underbrace{\mathrm{Memory}}_{\text{Context}}
\rightarrow
\underbrace{\mathrm{Planning}}_{\text{Decision}}
\rightarrow
\underbrace{\mathrm{Tool\ Use}}_{\text{Action}} \rightarrow
\underbrace{\mathrm{Observation}}_{\text{Feedback}}
\,\Bigr]_{\!n}
\rightarrow \mathrm{Solution}.
\end{aligned}
\end{equation*}
In each iteration $n$, the system must first retrieve relevant context from memory, reason over the current state to formulate a plan, execute a specific tool-incorporated action, and process the resulting observation. This cycle creates a compounding accumulation of tokens, where the output of step $n$ becomes the input cost of step $n+1$, resulting in high inference costs and slow response times. Consequently, mere model compression is insufficient. We therefore define an efficient agent as follows:
\begin{tcolorbox}[colframe=blue!75!black, colback=blue!10!white]
\textbf{Efficient agent} is not a smaller model, but as an agentic system optimized to maximize task success rates while minimizing resource consumption, including token usage, inference latency, and computational cost across memory, tool usage, and planning modules.
\end{tcolorbox}

Existing surveys have provided comprehensive views of individual agent components. Memory-oriented surveys~\cite{zhang2025survey,du2025rethinking,hu2026memoryageaiagents} mainly summarize memory taxonomies, storage and retrieval mechanisms, update operations, and memory evaluation. Tool use surveys~\cite{qu2025tool,xu2025llm} focus on how LLMs select, invoke, and learn to use external tools. Planning and reasoning surveys~\cite{huang2024understandingplanningllmagents,wei-etal-2025-plangenllms,zhao2025llm} organize studies on task decomposition, plan generation, reasoning frameworks, reflection, and decision-making. However, these component-centric perspectives do not directly answer a system-level efficiency question: under comparable task performance, how can an agent reduce token usage, latency, tool calls, planning steps, memory overhead, and computational cost?

Figure~\ref{fig:evolution} summarizes the emerging research landscape of efficient agents. Existing studies cluster around memory, tool use, planning, and benchmarks, suggesting that agent efficiency is not only a model-level problem but also a system-level problem across the agent workflow.

Our survey aims to systematize the numerous efforts in this emerging field from an efficiency-centered perspective. While efficient LLM surveys mainly focus on reducing the cost of the underlying model~\cite{llmsurvey1,llmsurvey2,llmsurvey3}, efficient agents require optimization at the system level. To bridge this gap, we categorize existing works into three strategic directions: 1) Efficient Memory: techniques for compressing historical context, managing memory storage, and optimizing context retrieval; 2) Efficient Tool Use: strategies to reduce unnecessary tool calls, lower external interaction latency, and improve tool-use decisions; and 3) Efficient Planning: strategies to reduce redundant reasoning steps, shorten execution trajectories, and decrease the number of API calls required to solve a problem.

It is important to note that these categories are analytical rather than mutually exclusive. In a deployed agent, memory, tool use, and planning are tightly coupled: memory can store tool-use experience and reusable plans; planning decides when to retrieve memory or invoke tools; tool observations become new memory and may reshape subsequent plans. When a method touches multiple components, we assign it to the category corresponding to its primary efficiency motivation or the main bottleneck it is designed to reduce. We then discuss cross-component effects throughout the paper, since a method's system-level efficiency often comes from how it changes costs in other modules rather than from its local module alone.

The remainder of this survey is organized as follows: Section~\ref{sec:preliminary} introduces the preliminaries and highlights the efficiency gap between agents and LLMs. Sections~\ref{sec:memory} through \ref{sec:efficient_planning} explore component-level efficiency, with a focus on memory, tool use, and planning optimizations. Subsequently, Section~\ref{sec:benchmark} addresses the quantification of efficiency. The survey concludes with a discussion on open challenges and future research directions.


\section{Preliminaries}
\label{sec:preliminary}
\subsection{Agent Formulation}

We model an LLM-based agent interacting with an environment as a partially observable
Markov decision process (POMDP) augmented with an external tool interface and an explicit memory component. Formally, we define the overall model as
\[
\mathcal{M} = (\mathcal{S}, \mathcal{O}, \mathcal{A}, P, R, \gamma;\ \mathcal{T}, \Psi;\ \mathcal{M}_{mem}, U, \rho).
\]
Here $\mathcal{S}$ denotes the latent environment state space, $\mathcal{O}$ the observation
space, and $\mathcal{A}$ the agent action space. The environment dynamics are given by the
transition kernel $P$, the reward function $R$, and the discount factor $\gamma\in[0,1)$.

The agent is additionally equipped with a set of external tools $\mathcal{T}$ and a tool
interface $\Psi$, which specifies how tool calls are executed and what tool outputs are
returned to the agent. Finally, we model explicit agent memory with memory state space
$\mathcal{M}_{mem}$, an update rule $U$ that maps the current memory and available information
to the next memory state, and an initialization distribution $\rho$ over the initial memory.

\subsection{From Pure LLMs to Agents}

\begin{wrapfigure}{r}{0.5\linewidth}
\vspace{-5pt}
  \centering
  \includegraphics[width=0.8\linewidth]{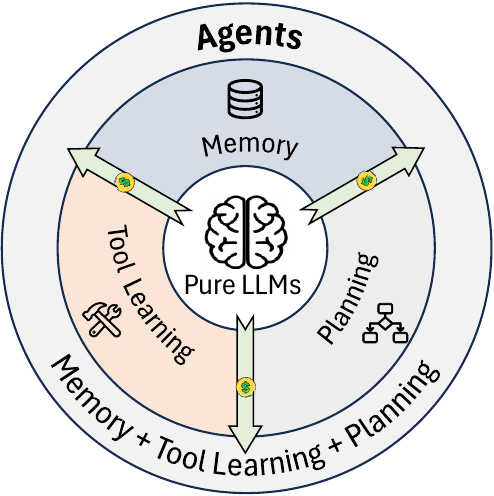}
  \caption{From LLMs to agents: standalone reasoning to trajectory-level reasoning with memory, planning, and tool use, while introducing additional cost sources.}
\vspace{-15pt}
  \label{fig:preliminary_fig}
\end{wrapfigure}

We define efficiency through a cost–performance trade-off: achieving comparable performance with lower cost, or achieving higher performance under a similar cost budget.

We acknowledge that many efficiency techniques used in LLM-based agents overlap with those for standalone LLMs (e.g., model compression and inference acceleration). In agents, however, these techniques mainly serve as foundational enablers rather than addressing the agent-specific sources of inefficiency. As summarized by \citet{wang2024survey}, compared to pure LLMs, LLM-based agents exhibit more human-like decision-making by augmenting a base model with cognitive components such as planning and memory.

Accordingly, in this subsection we focus on what differentiates agent efficiency from LLM efficiency. From a functional perspective, an agent is characterized by its ability to (i) plan and act over multiple steps, (ii) invoke external tools or environment commands to acquire information and execute operations, and (iii) condition subsequent decisions on retrieved or updated memory.

As illustrated in Figure~\ref{fig:preliminary_fig}, agentic systems introduce additional cost sources beyond generation. For a pure LLM, the inference cost is often dominated by token generation and can be approximated as:
\begin{equation*}
\mathrm{Cost}_{\text{LLM}} \approx \alpha\,N_{\text{tok}},
\end{equation*}
where $N_{\text{tok}}$ is the number of generated reasoning tokens and $\alpha$ captures
the per-token cost (e.g., time or monetary cost). In contrast, an agent may incur additional overhead from tools, memory, and retries as needed:
\begin{equation*}
\mathrm{Cost}_{\text{agent}}
\approx \alpha\,N_{\text{tok}}
+ \mathbb{I}_{\text{tool}}\cdot \mathrm{Cost}_{\text{tool}}
+ \mathbb{I}_{\text{mem}}\cdot \mathrm{Cost}_{\text{mem}}
+ \mathbb{I}_{\text{retry}}\cdot \mathrm{Cost}_{\text{retry}},
\end{equation*}
where $\mathbb{I}_{\text{tool}}, \mathbb{I}_{\text{mem}}, \mathbb{I}_{\text{retry}} \in \{0,1\}$ are indicator variables that equal $1$ if the agent invokes tools, accesses memory, or performs retries, respectively, and $0$ otherwise.
Therefore, improving agent efficiency is not only about reducing language generation, but also about reducing the frequency and improving the selectivity of tool or memory invocations and retries along a trajectory, to achieve a better cost–performance trade-off.
\section{Efficient Memory}
\label{sec:memory}
\begin{figure}[t]
    \centering
    \includegraphics[width=\linewidth]{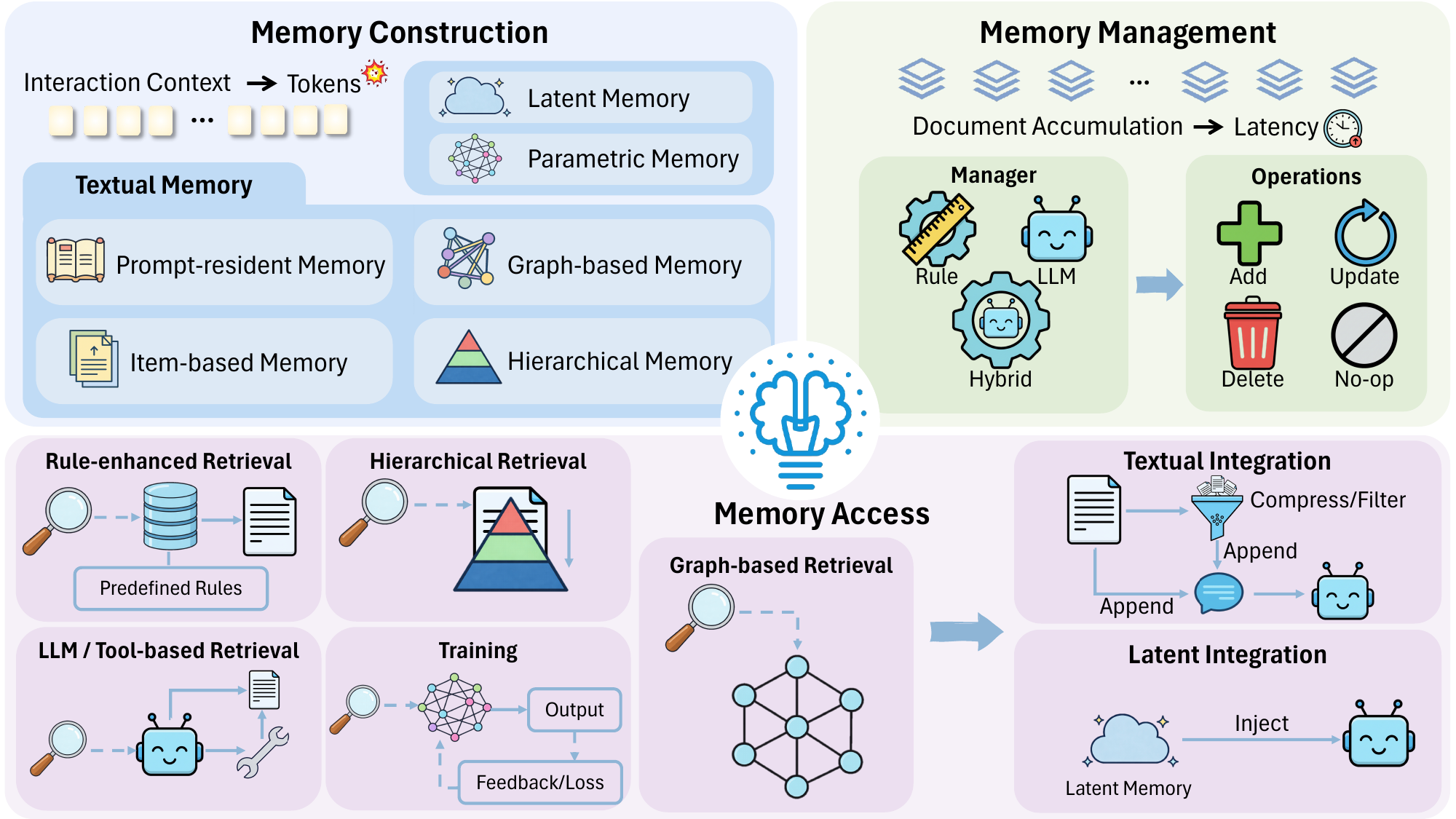}
    \caption{Efficient memory overview. This figure summarizes the agent-memory lifecycle in three phases: \textbf{Memory Construction}, which represents experience as textual memory or latent and parametric memory to mitigate token explosion and repeated context processing; \textbf{Memory Management}, which curates and updates an accumulating memory store via rule-based, LLM-based, or hybrid strategies to control latency; and \textbf{Memory Access}, which determines what memories to retrieve and how to integrate them into the model.}
    \label{fig:memory_picture_overview}
\end{figure}

A major efficiency bottleneck for LLM agents is the computational and token overhead induced by long contexts and long-horizon interactions, where agents may repeatedly reprocess large histories to act. \textbf{Memory-augmented reasoning} provides a principled way to alleviate this inefficiency. By storing and reusing past experience, including successes, failures, and interaction traces, agents can avoid redundant computation, make more informed decisions, and reduce costly retries. In this sense, memory is not merely an auxiliary component. It is a key mechanism for improving the overall efficiency-effectiveness trade-off of agent systems.

Our taxonomy is mainly inspired by the memory-operation view in \citet{du2025rethinking}, which summarizes six basic memory operations. However, in practical agent systems, these operations are often not performed in isolation. For example, memory updating and forgetting may be jointly decided within the same memory pipeline. Therefore, instead of treating individual operations as mutually exclusive categories, we organize this section from a lifecycle and pipeline perspective, covering \textbf{memory construction}, \textbf{memory management}, and \textbf{memory access}. This organization allows us to examine the efficiency implications of memory creation, maintenance, and retrieval, including their effects on token usage and latency.

For memory construction, we draw on the high-level distinction between parametric memory and contextual memory discussed by \citet{du2025rethinking}. Since our focus is on how agent memories are constructed and represented, we divide memory construction into \textbf{textual memory} and \textbf{latent and parametric memory}. Textual memory refers to explicit, context-accessible records, such as summaries, trajectories, reflections, and external memory entries. In contrast, latent and parametric memory stores experience in implicit representations or model parameters, rather than as directly readable text. The notion of latent memory has also been recognized in recent discussions of AI-agent memory~\citep{hu2026memoryageaiagents}, which further supports the need to include this category.

Because memory is central to efficiency gains, how to design an efficient memory module becomes an important problem. We therefore discuss efficiency-oriented designs throughout this lifecycle, focusing on how to maximize the benefit of memory while minimizing additional overhead. Figure~\ref{fig:memory_picture_overview} provides a structured overview of our taxonomy, and Tables~\ref{tab:efficient_memory_summary} and~\ref{tab:efficient_memory_summary_part2} summarize all memory-related methods discussed in this section for an at-a-glance overview.

\begin{table*}[!t]
\centering
\caption{Memory overview of efficiency-oriented mechanisms, Part I. Together with Table~\ref{tab:efficient_memory_summary_part2}, this table covers all memory-related methods discussed in this section. Part I includes latent and parametric memory and textual memory.}

\label{tab:efficient_memory_summary}
\renewcommand{\arraystretch}{0.72}
\setlength{\tabcolsep}{3pt}
\resizebox{\textwidth}{!}{%
\begin{tabular}{l|l|l|l}
\toprule
\textbf{Method} & \textbf{Category} & \textbf{Core Mechanism} & \textbf{Resource Link} \\
\midrule
\multicolumn{4}{c}{\textbf{\textit{Latent and Parametric Memory}}}\\
\midrule
Activation Beacon~\citep{zhang2024long} & Latent & Beacon-token KV compression for long context & \ghlink{https://github.com/FlagOpen/FlagEmbedding/}\\
MemoRAG~\citep{qian2025memorag} & Latent & Window-level memory tokens with updated KV cache & \ghlink{https://github.com/qhjqhj00/MemoRAG}\\
FlashMem~\citep{hou2026flashmemdistillingintrinsiclatent} & Latent & Shared-KV consolidation with entropy-triggered writes & N/A \\
Memory$^3$~\citep{yang2024memory3} & Latent & Sparse explicit KV knowledge injected into attention & N/A \\
MemGen~\citep{zhang2025memgen} & Latent & Triggered on-demand latent memory synthesis & \ghlink{https://github.com/KANABOON1/MemGen}\\
MemoryLLM~\citep{wang2024memoryllm} & Parametric & Fixed-size layer-wise memory pool with self-updates & \ghlink{https://github.com/wangyu-ustc/MemoryLLM}\\
M+~\citep{wang2025m+} & Parametric & GPU/CPU two-tier memory with co-trained retriever & \ghlink{https://github.com/wangyu-ustc/MemoryLLM}\\
Titans~\citep{behrouz2024titans} & Parametric & Test-time trainable neural memory with surprise-weighted updates & N/A \\

\midrule
\multicolumn{4}{c}{\textbf{\textit{Textual Memory}}}\\
\midrule
COMEDY~\citep{chen2025compress} & Prompt-resident & Two-stage memory distillation & \ghlink{https://github.com/nuochenpku/COMEDY}\\
 MemAgent~\citep{yu2025memagent} & Prompt-resident & Overwrite fixed memory & \ghlink{https://github.com/BytedTsinghua-SIA/MemAgent}\\
 MEM1~\citep{zhou2025mem1learningsynergizememory} & Prompt-resident & Update a compact shared internal state & \ghlink{https://github.com/MIT-MI/MEM1}\\
 AgentFold~\citep{ye2025agentfold} & Prompt-resident & Proactive context folding & N/A \\
 DC~\citep{suzgun2025dynamic} & Prompt-resident & Persistent, evolving memory & \ghlink{https://github.com/suzgunmirac/dynamic-cheatsheet}\\

MemoryBank~\citep{zhong2024memorybank} & Item-based & Ebbinghaus forgetting curve–based memory management & \ghlink{https://github.com/zhongwanjun/MemoryBank-SiliconFriend}\\
RECOMP~\citep{xu2023recomp} & Item-based & Compress retrieved documents & \ghlink{https://github.com/carriex/recomp}\\
Expel~\citep{zhao2024expel} & Item-based & experiential learning; insight distillation and management & \ghlink{https://github.com/LeapLabTHU/ExpeL}\\
Human-like memory~\citep{hou2024my} & Item-based & Cue-triggered memory recall & N/A \\
SeCom~\citep{pan2025secom} & Item-based & Segment-level memory; compression-based denoising for retrieval & \ghlink{https://github.com/microsoft/SeCom}\\
Memory-R1~\citep{yan2025memory} & Item-based & Adaptive memory CRUD and memory distillation, via two RL-trained agents & N/A \\
Mem0~\citep{chhikara2025mem0} & Item-based & Extract candidate memories; memory CRUD & \ghlink{https://github.com/mem0ai/mem0}\\
SimpleMem~\citep{liu2026simplememefficientlifelongmemory} & Item-based & Compact self-contained memories with query-adaptive retrieval and async consolidation & \ghlink{https://github.com/aiming-lab/SimpleMem}\\
agentic plan caching~\citep{zhang2025costefficientservingllmagents} & Item-based & Store plan template; plan cache lookup (hit/miss) and update & N/A \\
LD-Agent~\citep{li2025hello} & Item-based & Separate different memory; topic-based retrieval & \ghlink{https://github.com/leolee99/LD-Agent}\\
MemoChat~\citep{lu2023memochat} & Item-based & Structured on-the-fly memos & \ghlink{https://github.com/LuJunru/MemoChat}\\
RMM~\citep{tan2025prospect} & Item-based & Topic-based memory organization; consolidation (add/merge); online RL reranker & N/A \\
Memento~\citep{zhou2025memento} & Item-based & Parametric case retrieval via an online-updated Q-function & \ghlink{https://github.com/Agent-on-the-Fly/Memento}\\
MemInsight~\citep{salama2025meminsight} & Item-based & Attribute-augmented memory; attribute-guided retrieval & \ghlink{https://github.com/amazon-science/MemInsight}\\
ReasoningBank~\citep{ouyang2025reasoningbank} & Item-based & Distill strategies from failures and successes to cut exploration steps & N/A \\
A-MEM~\citep{xu2025mem} & Item-based & Atomic structured notes; link generation and memory evolution & \ghlink{https://github.com/agiresearch/A-mem}\\
ACE~\citep{zhang2025agentic} & Item-based & Incremental delta updates, lightweight merge and de-dup & \ghlink{https://github.com/ace-agent/ace}\\
Agent KB~\citep{tang2025agent} & Item-based & Cross-framework reusable experience Knowledge Base & \ghlink{https://github.com/OPPO-PersonalAI/Agent-KB}\\

GraphReader~\citep{li2024graphreader} & Graph-based & Graph-guided coarse-to-fine exploration & N/A \\
KG-Agent~\citep{jiang2025kg} & Graph-based & Tool-based hop-local KG processing & N/A \\
Zep~\citep{rasmussen2025zep} & Graph-based & Temporal KG memory & \ghlink{https://github.com/getzep/zep}\\
Mem0$^g$~\citep{chhikara2025mem0} & Graph-based & Extract candidate nodes; graph updation & \ghlink{https://github.com/mem0ai/mem0}\\
AriGraph~\citep{anokhin2024arigraph} & Graph-based & Memory graph; semantic-to-episodic cascading retrieval & \ghlink{https://github.com/AIRI-Institute/AriGraph}\\
D-SMART~\citep{lei2025d} & Graph-based & Structured OWL-compliant KG & N/A \\

MemGPT~\citep{packer2023memgpt} & Hierarchical & OS-style virtual memory paging for context & \weblink{https://research.memgpt.ai}\\
MemoryOS~\citep{kang2025memory} & Hierarchical & OS-inspired three-tier memory hierarchy with policy-based inter-tier updates & \ghlink{https://github.com/BAI-LAB/MemoryOS}\\
MemOS~\citep{li2025memos} & Hierarchical & Policy-guided type transformation of MemCubes across three memory forms & \ghlink{https://github.com/MemTensor/MemOS}\\
ReadAgent~\citep{lee2024human} & Hierarchical & Gist memory compression; on-demand lookup & \ghlink{https://github.com/read-agent/read-agent.github.io/blob/main/assets/read_agent_demo.ipynb}\\
HiAgent~\citep{hu2025hiagent} & Hierarchical & Subgoals as memory chunks; on-demand trajectory retrieval & \ghlink{https://github.com/HiAgent2024/HiAgent}\\
H-MEM~\citep{sun2025hierarchical} & Hierarchical & Layer-by-layer retrieval & N/A \\
LightMem~\citep{fang2025lightmem} & Hierarchical & Pre-compression; soft update (test-time); sleep-time update (offline) & \ghlink{https://github.com/zjunlp/LightMem}\\
xMemory~\citep{hu2026ragagentmemoryretrieval} & Hierarchical & Retrieval by decoupling query intent and aggregating multi-level memory evidence & \weblink{https://zhanghao-xmemory.github.io/Academic-project-page-template/}\\
HyMem~\citep{zhao2026hymemhybridmemoryarchitecture} & Hierarchical & Hybrid summary--raw memory with dynamic retrieval scheduling & \ghlink{https://github.com/xiaochenzhao-svg/HyMem} \\

\bottomrule
\end{tabular}%
}
\end{table*}

\begin{table*}[!t]
\centering
\caption{Memory overview of efficiency-oriented mechanisms, Part II. Together with Table~\ref{tab:efficient_memory_summary}, this table covers all memory-related methods discussed in this section. Part II includes procedural reuse via skills and multi-agent memory.}
\label{tab:efficient_memory_summary_part2}
\renewcommand{\arraystretch}{0.72}
\setlength{\tabcolsep}{3pt}
\resizebox{\textwidth}{!}{%
\begin{tabular}{l|l|l|l}
\toprule
\textbf{Method} & \textbf{Category} & \textbf{Core Mechanism} & \textbf{Resource Link} \\
\midrule
\multicolumn{4}{c}{\textbf{\textit{Procedural Reuse via Skills}}}\\
\midrule

Skill-Pro~\citep{mi2026skillprolearningreusableskills} & \textemdash{} & Non-parametric PPO for reusable skill extraction from experience & N/A \\
SkillRL~\citep{xia2026skillrlevolvingagentsrecursive} & \textemdash{} & Recursive reinforcement learning with skill library and policy co-improvement & \ghlink{https://github.com/aiming-lab/SkillRL}\\
MemSkill~\citep{zhang2026memskilllearningevolvingmemory} & \textemdash{} & Evolvable memory-operation skills for extraction, consolidation, and pruning & \weblink{https://viktoraxelsen.github.io/MemSkill/}\\
AutoSkill~\citep{yang2026autoskillexperiencedrivenlifelonglearning} & \textemdash{} & Experience-driven lifelong skill self-evolution from user interactions & \ghlink{https://github.com/ECNU-ICALK/AutoSkill}\\
Trace2Skill~\citep{ni2026trace2skilldistilltrajectorylocallessons} & \textemdash{} & Distill trajectory-local lessons into transferable agent skills & \ghlink{https://github.com/Qwen-Applications/Trace2Skill}\\
CoEvoSkills~\citep{zhang2026coevoskillsselfevolvingagentskills} & \textemdash{} & Co-evolutionary verification for self-evolving skill libraries & \weblink{https://zhang-henry.github.io/CoEvoSkills/}\\
SKILLFOUNDRY~\citep{shen2026skillfoundrybuildingselfevolvingagent} & \textemdash{} & Build self-evolving skill libraries from heterogeneous scientific resources & \weblink{https://ma-compbio-lab.github.io/SkillFoundry/\#overview}\\
Graph-of-Skills~\citep{liu2026graphofskillsdependencyawarestructuralretrieval} & \textemdash{} & Dependency-aware graph retrieval over massive agent skill libraries & \ghlink{https://github.com/davidliuk/graph-of-skills}\\
SkillClaw~\citep{ma2026skillclawletskillsevolve} & \textemdash{} & Collective skill evolution with an agentic evolver & \ghlink{https://github.com/AMAP-ML/SkillClaw}\\
SkillOS~\citep{ouyang2026skilloslearningskillcuration} & \textemdash{} & Learn skill curation policies for maintaining self-evolving skill repositories & N/A \\
SkillLens~\citep{miao2026skilllensadaptivemultigranularityskill} & \textemdash{} & Adaptive multi-granularity skill reuse to avoid loading or rewriting whole skills & N/A \\

\midrule
\multicolumn{4}{c}{\textbf{\textit{Multi-Agent Memory}}}\\
\midrule

MS~\citep{gao2024memory} & Shared & Shared memory pool; selective addition; continual retriever training & \ghlink{https://github.com/GHupppp/InteractiveMemorySharingLLM}\\
G-Memory~\citep{zhang2025g} & Shared & three-tier graph memory with bi-directional coarse-to-fine retrieval & \ghlink{https://github.com/bingreeky/GMemory}\\
RCR-Router~\citep{liu2025rcr} & Shared & Feedback-refined iterative router under a token budget & N/A \\
MemIndex~\citep{saleh2025memindex} & Shared & Intent-indexed bipartite graphs; semantic slicing and dynamic indexing & N/A \\
MIRIX~\citep{wang2025mirix} & Shared & Six-module hierarchical memory with staged retrieval and parallel updates & \ghlink{https://github.com/Mirix-AI/MIRIX}\\
Intrinsic Memory Agents~\citep{yuen2025intrinsic} & Local & Role-aligned templates; intrinsic iterative updates & N/A \\
AgentNet~\citep{yang2025agentnet} & Local & Fixed-size memory modules for routing/execution; dynamic pruning & \ghlink{https://github.com/zoe-yyx/AgentNet}\\
DAMCS~\citep{yang2025llm} & Local & Decentralized per-agent STWM/LTM with goal-oriented hierarchical knowledge graph & \weblink{https://happyeureka.github.io/damcs/}\\
SRMT~\citep{sagirova2025srmt} & Mixed & Personal latent memory and globally broadcast shared recurrent memory & \ghlink{https://github.com/Aloriosa/srmt}\\
Collaborative Memory~\citep{rezazadeh2025collaborative} & Mixed & Policy-based filtering/transformation of memory fragments; shared-memory reuse & N/A \\
LEGOMem~\citep{han2025legomem} & Mixed & Role-aware memory routing; runtime-efficient retrieval scheduling & N/A \\

\bottomrule
\end{tabular}%
}
\end{table*}




\subsection{Memory Construction}
Memory construction determines how raw interaction history, documents, observations, and execution traces are converted into reusable memory representations. From an efficiency perspective, this step is crucial because the initial representation largely determines later costs: a verbose memory increases attention and retrieval overhead, whereas an overly compressed memory may omit information needed for correct decisions. Therefore, efficient construction is not simply about storing more experience, but about transforming experience into compact, accessible, and task-relevant forms.

Following the distinction introduced above, we discuss two complementary construction paradigms. \textbf{Latent and parametric memory} compresses experience into hidden states, KV caches, memory tokens, or model-side parameters to reduce repeated context processing. \textbf{Textual memory} keeps information in readable natural-language or structured text, making it easier to inspect, edit, retrieve, and cite. Both paradigms aim to mitigate token growth and long-context degradation, such as the “lost in the middle” phenomenon \citep{liu2024lost}, but they make different efficiency trade-offs between latency, interpretability, storage cost, and retrieval flexibility.



\subsubsection{Latent and Parametric Memory}
Latent and parametric memory store agent experience in non-textual forms that directly condition model computation, avoiding the cost of serializing and re-reading full interaction histories as prompt tokens. The two types differ fundamentally in \emph{where} information is stored. \textbf{Latent memory} retains experience in activation space without modifying any model parameters. \textbf{Parametric memory} encodes information into the \emph{parameters} of a dedicated memory module.

On the latent side, one group of methods builds compact representations by compressing long contexts into a small set of KV activations.
Activation Beacon~\citep{zhang2024long} interleaves special beacon tokens into chunked input and uses progressive layer-wise compression to distill raw-token KV activations into the beacons, which accumulate as latent memory while raw activations are discarded.
MemoRAG~\citep{qian2025memorag} inserts memory tokens after each context window and updates their KV cache across windows with separate weight matrices, yielding a compact global memory that can later serve as retrieval clues for downstream generation.
FlashMem~\citep{hou2026flashmemdistillingintrinsiclatent} avoids auxiliary encoders by treating the last hidden state as a sufficient statistic for interaction history; a Shared-KV Consolidator attends directly to the backbone's frozen KV cache to synthesize compact memory, and a parameter-free Cognitive Monitor triggers consolidation only under high attention entropy.
Memory$^3$~\citep{yang2024memory3} takes a complementary approach: rather than accumulating episodic experience, it converts a static knowledge base into sparse explicit key-value pairs injected into self-attention layers at decoding time, treating externalized knowledge as a middle ground between plain-text RAG and full model parameters in terms of read/write cost.
MemGen~\citep{zhang2025memgen} generates latent memory on demand within the reasoning stream: an RL-trained memory trigger monitors hidden states to decide when memory invocation is needed, and a memory weaver produces a compact latent token sequence encapsulating relevant experience that is injected back into ongoing reasoning; without explicit supervision, distinct latent clusters corresponding to planning, procedural, and working memory spontaneously emerge.

On the parametric side, methods encode experience into the parameters of a dedicated memory module.
MemoryLLM~\citep{wang2024memoryllm} augments a frozen backbone with an additional 1B-parameter memory pool of hidden vectors, one pool per transformer layer, which are concatenated into attention at each layer. A self-update mechanism modifies the pool in-place when new knowledge arrives without backpropagation through the backbone, while a random-dropping rule exponentially phases out old entries to maintain fixed capacity.
M+~\citep{wang2025m+} extends MemoryLLM by adding a GPU/CPU two-tier long-term memory alongside the short-term pool; a co-trained retriever fetches a small number of relevant hidden-state vectors per layer from the CPU-resident store, pushing effective knowledge retention from 20K to over 160K tokens without increasing GPU memory.
Titans~\citep{behrouz2024titans} takes a more explicitly gradient-based approach: a standalone neural memory module updates its own weights at test time via a surprise-weighted gradient signal, writing strongly only when prediction error is high and applying a forgetting decay proportional to memory capacity. Because history is encoded into the module's parameters through actual weight updates, Titans can be understood as a fast-weights or meta in-context learner that operates alongside a conventional attention head.

Across both categories, a common design principle emerges: memory writes are selectively gated so that the agent avoids redundantly re-encoding stable or low-information content. The key practical trade-off is that latent memory updates are fast and require no gradient computation but are bounded by cache or context capacity, whereas parametric memory can in principle encode arbitrary associations with no hard capacity limit at the cost of a forward-backward pass or an equivalent weight-update step per memory write.

\begin{summarybox}[]{Latent and Parametric Memory}
\begin{itemize}[leftmargin=0.5em]
\item \textbf{Pros}: Compresses long interaction histories into compact hidden states or KV caches, allowing agents to reuse prior context without repeatedly replaying full textual trajectories.
\item \textbf{Cons}: The stored information is less interpretable than text and may require additional training, cache management, or memory-token infrastructure; aggressive compression can also discard details needed for downstream decisions.
\item \textbf{Suitable situations}: Best for efficiency-critical, long-horizon agents where historical state must be reused frequently and the main bottleneck is repeated prefill, long-context attention, or token-level communication.
\item \textbf{Trade-offs}: Compared with textual memory, latent and parametric memory save tokens and reduce latency but sacrifice transparency and editability; parametric approaches additionally incur a per-write update cost that latent methods avoid.
\end{itemize}
\end{summarybox}
\subsubsection{Textual Memory}
\label{sec:external_memory}
Textual memory represents past experience as explicit text or structured symbolic records, including summaries, dialogue segments, trajectories, profiles, plans, and graph facts. Compared with latent memory, it is more transparent, but it can still become inefficient if the stored text is too long, redundant, or poorly indexed. Thus, the central efficiency question for textual memory is how to preserve useful information while controlling prompt length, retrieval latency, and maintenance overhead.
\paragraph{Prompt-resident Memory.}
Prompt-resident memory keeps a compact textual state directly inside the context window. Its efficiency goal is to retain immediately useful information while preventing the prompt from growing with the full interaction history. In practice, this is often achieved by frequently \textbf{rewriting or compressing} the memory as the process evolves.

COMEDY~\citep{chen2025compress} uses an LLM to generate and compress memory: it extracts session-specific memories from past conversations and then condenses them into a compact representation of key events, the user profile, and relationship changes.
MemAgent~\citep{yu2025memagent} and MEM1~\citep{zhou2025mem1learningsynergizememory} both process long inputs sequentially by rewriting and updating a compact memory state at each step: MemAgent updates a summarized memory after each chunk, while MEM1 uses reinforcement learning \cite{zhang2025surveyrein} to maintain a fixed-length internal state tagged by <IS></IS> that replaces itself in the next prompt.
AgentFold~\citep{ye2025agentfold} proactively folds interaction history into multi-scale summaries plus the latest full turn, slowing critical information loss while reducing token usage.

By retaining a compact memory in the prompt rather than the full history, these methods reduce the effective context length the LLM needs to attend to, thereby improving long-context performance while decreasing computational cost and increasing efficiency.

\paragraph{Item-based Memory.}
Item-based memory stores experience as discrete textual units, such as events, insights, profiles, or plan templates. Its efficiency depends on whether each item is compact enough to retrieve and insert cheaply while still preserving the information needed for future decisions. Early agent-memory systems often store full trajectories or experiences, sometimes alongside summaries, which can lead to long context and inefficient retrieval.
MemoryBank~\citep{zhong2024memorybank} stores daily conversation records and summarizes past events and user profiles from these conversations, but it incurs high token costs. Similarly, Expel~\citep{zhao2024expel} suffers from a similar limitation, as it accumulates experiences through trial and error and distills them into natural-language insights.

To improve efficiency, many works use \textbf{memory extraction, compression, or summarization} to convert raw interactions into shorter and more informative memory items. This reduces input token consumption while making each stored entry easier to retrieve and reuse.
Human-like memory~\citep{hou2024my} extracts episodic memories from users dialogues, encapsulating content and temporal context into database structure. SeCom~\citep{pan2025secom} uses segmentation model to divide long-term conversations into topic-coherent segmentation, and applies the compress model to denoise the segmentation, which further promotes efficient retrieval.
Memory-R1~\citep{yan2025memory} and Mem0~\citep{chhikara2025mem0} both extract and summarize ongoing dialogue into candidate memories for downstream updating; Memory-R1 does so at each turn, while Mem0  forms candidate memories from the new message pair, using a conversation summary and recent messages as context.
Agentic plan caching~\citep{zhang2025costefficientservingllmagents} turns a successful run into a reusable cache entry by rule-based filtering the execution log and then using a lightweight LLM to remove context-specific details, storing the result as a (keyword, plan template) pair. LD-Agent~\citep{li2025hello} separates event and persona memory, using short-term and long-term banks for timestamped dialogue context and embedded event summaries, and a persona extractor to store user and agent traits in long-term persona banks.
SimpleMem~\citep{liu2026simplememefficientlifelongmemory} follows a compression-oriented design: it filters redundant dialogue content, resolves context-dependent references, and rewrites useful information into compact, self-contained memory units with multi-view indices.


Beyond extraction and compression, another way to improve efficiency is to \textbf{design more structured memory systems}. Structure reduces retrieval ambiguity by attaching useful indices, attributes, or reusable abstractions to each memory item, enabling faster lookup and better use of stored information.
One representative structure is topic-indexed memory, which organizes interactions into topic-level groups and stores each topic summary together with its corresponding dialogue segment for efficient retrieval.
MemoChat~\citep{lu2023memochat} and RMM~\citep{tan2025prospect} both build topic-indexed memories. MemoChat records topic–summary–dialogue entries on the fly, while RMM groups each session by topic and stores each topic summary with its corresponding dialogue segment.
These methods can be viewed as enriched-item construction, where raw interactions are transformed into textual memory units equipped with additional annotations for later retrieval and association. MemInsight~\citep{salama2025meminsight} annotates memories with LLM-mined attribute--value pairs, while A-MEM~\citep{xu2025mem} converts each interaction into an atomic note with LLM-generated contextual descriptions, keywords, and tags. MemGAS~\citep{xu2026from} follows the same paradigm at the session level by augmenting sessions with turn-level dialogues, LLM-generated summaries, and keywords, and further links new memories to relevant historical memories through GMM-based clustering.
A further strategy is to distill experience libraries for reusable decision-making, summarizing trajectories or execution logs into standardized entries that capture reusable strategies, domain concepts, and common failure modes.
ReasoningBank~\citep{ouyang2025reasoningbank} summarizes successful and failed trajectories into structured memory items with a title, brief description, and content, and stores them with the task query and trajectory for embedding-based retrieval.
ACE~\citep{zhang2025agentic} represents context as structured, itemized bullets, each with a unique identifier, counters tracking how often it was marked helpful or harmful, and content such as a reusable strategy, domain concept, or common failure mode.
Agent KB~\citep{tang2025agent} turns execution logs into structured experience entries through human-guided abstraction, using few-shot prompting and a standardized cross-framework action vocabulary.

\paragraph{Graph-based Memory.} 
Graph-based memory is a more structured form of textual memory that represents entities, events, and relations as nodes and edges. This structure improves efficiency by merging repeated information about the same entity, localizing relevant facts through graph traversal, and supporting multi-hop evidence access without replaying the entire history.
Some methods focus on constructing graph-structured representations from long inputs or knowledge-graph interactions so that multi-hop evidence can be organized and accessed efficiently.
Targeted at the long context task, GraphReader~\citep{li2024graphreader} segments long text into chunks, compresses them into key elements and atomic facts, and uses these to construct a graph that captures long-range dependencies and multi-hop relations. KG-Agent~\citep{jiang2025kg} constructs a task-specific subgraph by tool calls and records the retrieved entities and relations as knowledge memory.

Another line of work constructs long-term memory directly as a dynamic knowledge graph, turning interactions into entities, relations, and time-aware facts that can be incrementally updated.
Zep~\citep{rasmussen2025zep} builds memory as a temporally aware knowledge graph by ingesting time-stamped episodes, extracting and aligning entities and relations, and storing fact edges with periods of validity. It also constructs a community subgraph that clusters strongly connected entities and stores high-level community summaries.
Mem0$^g$~\citep{chhikara2025mem0} represents memory as a directed labeled graph, where an LLM converts new messages into entities and relation triplets that form candidate nodes or edges for graph updates.
D-SMART~\citep{lei2025d} incrementally constructs an OWL-compliant dialogue KG by first distilling each turn into an assertion-like statement, then converting it into a KG fragment for integration.
AriGraph~\citep{anokhin2024arigraph} updates a unified semantic–episodic memory graph online by adding an episodic node for each observation and extracting triplets to update the semantic graph, linking the two via episodic edges.
MAGMA~\citep{jiang2026magma} represents agent memory as a time-variant multi-graph, where each memory item is organized through four orthogonal relational graphs: semantic, temporal, causal, and entity graphs.

Graph-based memory represents entities and their relations as a structured graph. Building the graph already compresses and normalizes the history by merging repeated content about the same entity into a single node and keeping only relevant relations as edges. This makes construction more efficient by producing a compact structure that avoids unbounded prompt growth and supports fast retrieval later.

\paragraph{Hierarchical Memory.}
Hierarchical memory organizes information into multiple linked levels, enabling coarse-to-fine, on-demand access. Most hierarchical memory methods consider both structure and content, but with different emphases. Accordingly, related work can be grouped by whether it places more weight on structural organization and management or on content abstraction and indexing.

\textbf{System-oriented} hierarchical memory designs define explicit storage tiers and interfaces for reading and updating memory to manage long interaction history.
MemGPT~\citep{packer2023memgpt} constructs a hierarchical memory by partitioning the in-context prompt into system instructions, a writable working context, and a FIFO message buffer, and storing the remaining history and documents in external recall and archival memory.
MemoryOS~\citep{kang2025memory} adopts an OS-inspired hierarchical memory design with three storage tiers: short-term memory stores recent dialogue pages, mid-term memory groups pages into topic segments with summaries, and long-term personal memory maintains user and agent persona information.
MemOS~\citep{li2025memos} standardizes memory as MemCubes, each composed of a structured metadata header and a memory payload that can encapsulate plaintext, activation states, or parameter deltas. New interactions are incrementally turned into MemCubes and organized in a hierarchical structure.

\textbf{Content-oriented} approaches build hierarchical indices by segmenting and compressing documents or trajectories into multi-granularity summaries.
ReadAgent~\citep{lee2024human} splits a long document into pages and summarizes each page into a page-linked gist memory, forming a simple hierarchical index.
HiAgent~\citep{hu2025hiagent} compresses working memory into subgoals and observations, and stores full trajectories in external memory indexed by these summaries.
H-MEM~\citep{sun2025hierarchical} constructs a hierarchical structure, with four memory layers:  Domain Layer, Category Layer, Memory Trace Layer, and Episode Layer. It designs prompts to guide model to parse the interactions into these layers, which forms a progressively optimized index. LightMem~\citep{fang2025lightmem} uses a sensory–STM–LTM pipeline that first pre-compresses inputs by the sensory module, then groups turns into topic segments for STM and periodically summarizes these segments into compact LTM entries.
HyMem~\citep{zhao2026hymemhybridmemoryarchitecture} also uses event-level organization for long dialogues, but keeps summary memories linked to raw dialogue evidence, allowing compact access without fully discarding fine-grained details.

\begin{summarybox}[]{Textual Memory}
\begin{itemize}[leftmargin=0.5em]
\item \textbf{Pros}: Keeps memory human-readable, editable, and easy to cite or inspect, while moving most historical content outside the active prompt until it is needed.
\item \textbf{Cons}: Still incurs retrieval and reinsertion costs, and retrieved text can be noisy, redundant, or too long if the memory store is not well structured.
\item \textbf{Suitable situations}: Prompt-resident memory fits short, highly relevant state; item-based memory fits general long-trajectory agents; graph memory fits entity--relation and multi-hop reasoning; hierarchical memory fits ultra-long histories or large corpora needing coarse-to-fine access.
\item \textbf{Trade-offs}: More structure improves retrieval precision and scalability, but increases construction and maintenance overhead.
\end{itemize}
\end{summarybox}

\subsection{Memory Management}
After memories are constructed, their efficiency depends on how they are maintained over time. If every new observation is simply appended, the memory store quickly becomes redundant, stale, and expensive to retrieve from; retrieval latency increases, prompt budgets are wasted on low-value content, and noisy memories can harm downstream decisions. Memory management therefore controls the lifecycle of memory through operations such as updating, merging, deleting, consolidating, and archiving.

We group management strategies by the cost and adaptivity of their decision process. \textbf{Rule-based management} uses lightweight signals such as recency, frequency, capacity, or explicit feedback. \textbf{LLM-based management} makes more semantic decisions about what should be stored or revised, but incurs additional model calls. \textbf{Hybrid management} combines the two, using cheap rules for routine maintenance and invoking LLMs only when semantic judgment is likely to improve memory quality.

\subsubsection{Rule-based Management}
Rule-based management applies predefined policies to update, remove, or merge memories. Its main appeal is efficiency: decisions can be made without extra LLM calls, so memory size and retrieval cost remain bounded even during long interactions. The limitation is that such rules are usually only indirect proxies for usefulness, which means they may discard semantically important but infrequent information.

MemoryBank~\citep{zhong2024memorybank} introduces an Ebbinghaus-inspired memory update rule that decays memories over time while reinforcing important ones. Building on this idea, H-MEM~\citep{sun2025hierarchical} retains forgetting-curve-based decay and further adds feedback-driven regulation to dynamically adjust memory according to user feedback. Experimental results in A-MEM~\citep{xu2025mem} suggest that forgetting-curve-based memory management effectively controls memory size and reduces retrieval time. However, it also leads to a substantial drop in task performance.

Apart from forgetting-curve-based policies, a common rule-based strategy is trigger-driven memory maintenance, such as evicting or migrating items when a fixed-size buffer reaches capacity (e.g., FIFO replacement) \citep{packer2023memgpt,kang2025memory}.
In practice, these simple rules are often intertwined with LLM-based management, where the model summarizes or saves key information before items are removed or moved; more details are discussed in Section~\ref{sec:hybrid-management}.
\begin{summarybox}[]{Rule-based Management}
\begin{itemize}[leftmargin=0.5em]
\item \textbf{Pros}: Fast, predictable, and inexpensive because updates, pruning, and merging follow predefined rules without extra LLM calls.
\item \textbf{Cons}: Static rules are often task-agnostic, so they may remove rare but critical memories or retain frequent but low-value ones.
\item \textbf{Suitable situations}: Works well when memory importance can be approximated by simple signals such as recency, frequency, capacity, or explicit user feedback.
\item \textbf{Trade-offs}: Minimizes management overhead but gives up semantic adaptivity.
\end{itemize}
\end{summarybox}

\subsubsection{LLM-based Management}
LLM-based management uses the model itself to judge memory importance, consistency, and redundancy. Compared with fixed rules, it can make more task-aware decisions about whether to add, update, merge, or delete memory entries. This improves the relevance and compactness of the memory store, but the additional LLM calls also increase latency and monetary cost. Existing methods can be broadly categorized by their decision form: selecting from a discrete set of operations versus generating open-ended updates.

A common formulation is operation selection, where the model \textbf{picks an action from a predefined set}, such as ADD, DELETE, or UPDATE, and applies it to retrieved memories.
Both Memory-R1~\citep{yan2025memory} and Mem0~\citep{chhikara2025mem0} update an external memory by retrieving similar entries and choosing among {ADD, UPDATE, DELETE, NOOP}. Memory-R1 learns the choice via reinforcement learning, while Mem0 lets an LLM select the operation after vector-based retrieval. RMM~\citep{tan2025prospect} follows the same retrieve-then-update pattern: for each newly extracted topic memory, it retrieves the top-$k$ most similar entries from the memory bank and prompts an LLM to decide whether to merge or add. Separately, ExpeL~\citep{zhao2024expel} maintains an insights list through direct list editing, applying operations such as ADD, EDIT, UPVOTE, and DOWNVOTE to correct or gradually suppress erroneous and outdated insights.

A different formulation casts memory management as \textbf{open-ended generation}, where the model produces the update itself and implicitly performs the update operation rather than picking from a fixed action set. A-MEM~\citep{xu2025mem} uses generative updates: it retrieves top-$k$ similar notes with a fixed encoder, then an LLM creates links and rewrites related notes via memory evolution.

\begin{summarybox}[]{LLM-based Management}
\begin{itemize}[leftmargin=0.5em]
\item \textbf{Pros}: Makes semantic, task-aware decisions about what to write, merge, summarize, or forget, often producing cleaner and more useful memory than fixed rules.
\item \textbf{Cons}: Adds LLM calls during memory management, increasing latency, monetary cost, and potential error propagation from imperfect judgments.
\item \textbf{Suitable situations}: Useful when memory quality matters more than update speed, especially for open-ended tasks where importance cannot be captured by simple heuristics.
\item \textbf{Trade-offs}: Improves relevance and compression quality at the cost of higher online computation.
\end{itemize}
\end{summarybox}

\subsubsection{Hybrid Management}
\label{sec:hybrid-management}
Hybrid management aims to obtain the efficiency of rules and the semantic awareness of LLM-based updates. In practice, many systems use inexpensive triggers for frequent operations, such as buffer overflow, deduplication, or tier transfer, while reserving LLM calls for summarization, conflict resolution, or consolidation. This design is particularly important for deployed agents, where memory is updated continuously and fully LLM-based maintenance would make every interaction more expensive. 

Typical designs include tier-specific management, where rule-based triggers promote or consolidate information across tiers and costly LLM updates are invoked only when necessary.
MemoryOS~\citep{kang2025memory} and LightMem~\citep{fang2025lightmem} both adopt tier-specific, trigger-driven updates for hierarchical memory. MemoryOS manages STM as FIFO pages with overflow migrated to MTM, uses segment Heat scores in MTM for eviction and promotion, and updates LPM via an LLM, whereas LightMem triggers topic segmentation when the sensory buffer is full, summarizes topics into LTM when STM exceeds a token budget, and combines online soft updates with offline sleep-time consolidation. LD-Agent~\citep{li2025hello} uses a time-gap threshold as the trigger, summarizing the short-term cache into a long-term event record and clearing the cache to mark session boundaries.
MemGPT~\citep{packer2023memgpt} uses a hierarchical memory with main context and external context. A Queue Manager enforces token limits via memory pressure warnings, eviction, and recursive summarization, while a Function Executor turns model outputs into function calls to read and write across tiers.

Another management is item-level selection and pruning, using rules or heuristics for fast de-duplication and removal while relying on LLMs for semantic keep-or-drop decisions.
Agent KB~\citep{tang2025agent} and ACE~\citep{zhang2025agentic} exemplify item-level selection and pruning for hybrid memory management. Agent KB reduces redundancy by thresholding embedding similarity and using an LLM ranker to keep the better experience, then evicts low-utility entries based on a learned utility score. ACE maintains a bulletized context through incremental delta updates and applies embedding-based grow-and-refine to merge, prune, and de-duplicate bullets, keeping the context compact.

Besides, some management also considers lifecycle policies that use lightweight metrics to schedule costly maintenance beyond tier transfer, such as consolidation, deduplication, and archiving.
MemOS~\citep{li2025memos} manages MemCubes with explicit lifecycle and version tracking, using policy- and metric-driven modules such as MemScheduler and MemVault for deduplication, conflict handling, and archiving. Crucially, it supports type-aware transformation across Plaintext Memory, Activation Memory, and Parameter Memory, including promotion and demotion between types.
SimpleMem~\citep{liu2026simplememefficientlifelongmemory} complements this lifecycle view by asynchronously consolidating related compact memory units into higher-level abstractions, reducing semantic redundancy while moving part of the maintenance cost away from online interaction.

For graph-structured memory, hybrid management applies rule-based graph updates, while using LLMs to retrieve relevant subgraphs and verify contradictions or outdated content before updating relations.
Zep~\citep{rasmussen2025zep}, Mem0$^{g}$~\citep{chhikara2025mem0}, and AriGraph~\citep{anokhin2024arigraph} follow a similar pattern for graph memory maintenance: an LLM judges semantic conflicts or staleness against retrieved related edges, while the graph is updated through rule-based operations such as edge invalidation or removal and insertion of new relations to preserve temporal or world-model consistency. Additionally, D-SMART~\citep{lei2025d} maintains an OWL-compliant Dynamic Structured Memory and performs two-stage conflict resolution by letting an LLM identify contradicted or superseded triples, pruning them before merging the new fragment, with an optional OWL reasoner for logical consistency checking.
MAGMA~\citep{jiang2026magma} retrieves memories via policy-guided graph traversal. It identifies multi-signal anchor nodes through hybrid retrieval with RRF, expands them using intent-aware heuristic beam search, and serializes the visited nodes into a narrative context.

\begin{summarybox}[]{Hybrid Management}
\begin{itemize}[leftmargin=0.5em]
\item \textbf{Pros}: Combines cheap rule-based control with selective LLM judgment, so routine updates remain lightweight while difficult cases receive semantic processing.
\item \textbf{Cons}: Introduces more design complexity, including when to trigger LLM management, how to synchronize tiers, and how to avoid conflicting policies.
\item \textbf{Suitable situations}: Appropriate for deployed agents with frequent interactions, where most memory operations should be cheap but some consolidation or conflict resolution requires deeper reasoning.
\item \textbf{Trade-offs}: Often offers the best efficiency--quality balance, but only if trigger policies are well calibrated.
\end{itemize}
\end{summarybox}
\subsection{Memory Access}
Memory access retrieves and uses a small subset of a large memory bank for the current query or state. It directly affects retrieval latency, inserted-token cost, and downstream generation quality. Efficient access must avoid two opposite failures: missing a critical memory, which may cause wrong actions or repeated exploration, and retrieving too much, which reintroduces context bloat.

\subsubsection{Memory Selection}
Memory selection determines which memories should be recalled for the current query or state. 
From an efficiency perspective, the goal is to maximize the usefulness of retrieved memories while reducing retrieval latency, context length, and downstream reasoning overhead. 
Although many methods do not impose explicit budgets during selection, they are often evaluated by efficiency-related metrics such as token usage, retrieval time, or performance under limited context. 
Most methods follow vanilla retrieval, i.e., encoding the query and its context into embeddings and selecting relevant information via similarity search, while others employ improved retrieval mechanisms to enhance retrieval quality and reduce redundant context.

\paragraph{Rule-enhanced Retrieval.}
Some methods enhance retrieval by incorporating additional rule-based scoring factors and applying preprocessing steps before retrieval. Generative Agents~\citep{park2023generative} and Human-like memory~\citep{hou2024my} take time into consideration, namely recency and elapsed time in these works. Apart from this, Generative Agents adds importance, a score generated by LLM based on the semantic importance, and Human-like memory adds recall frequency, computed according to the mathematical model. Agent KB~\citep{tang2025agent} employs a hybrid retrieval strategy that integrates lexical matching with semantic ranking by task similarity, combining both signals into a unified retrieval score. For long-term event retrieval, LD-Agent~\citep{li2025hello} combines semantic relevance, noun-based topic overlap, and an exponential time-decay factor into an overall score, and only retrieves memories whose semantic similarity exceeds a threshold.
While the aforementioned methods improve retrieval by adding additional scoring factors while keeping the computational cost comparable to vanilla retrieval, MemInsight~\citep{salama2025meminsight} augments memories with LLM generated attribute value annotations and leverages these augmentations for retrieval, either by filtering memories via attribute matching or by embedding the aggregated augmentations for vector similarity search. 
MemGAS~\citep{xu2026from} applies LLM-based filtering to the retrieved multi-granularity memories, removing redundant and noisy content before response generation.
RF-Mem~\citep{zhang2026evoking} improves retrieval efficiency with a familiarity uncertainty-guided dual-path retriever. It routes high-familiarity queries to a fast top-$K$ Familiarity path, while activating a deeper Recollection path for uncertain queries through clustering and iterative evidence expansion in embedding space.
SimpleMem~\citep{liu2026simplememefficientlifelongmemory} similarly controls online access cost through query-adaptive retrieval, using a small set of compact memories for simple queries and expanding to broader consolidated memories only when complex multi-fact reasoning is needed.

\paragraph{Graph-based Retrieval.}
For graph-based memory, retrieval naturally follows the graph structure, enabling efficient neighbor expansion and more precise localization of relevant facts, especially when queries target entity- and relation-centric information.
Given a textual query, Both AriGraph~\citep{anokhin2024arigraph} and Mem0$^g$~\citep{chhikara2025mem0} retrieve from a memory graph by anchoring on query-relevant facts and expanding neighbors into a local subgraph. AriGraph retrieves semantic triplets and then ranks episodic vertices via episodic search, whereas Mem0$^g$ pairs entity-centric subgraph construction with semantic triplet retrieval over relationship triplets.
MAGMA~\citep{jiang2026magma} retrieves memories via policy-guided graph traversal. It identifies multi-signal anchor nodes through hybrid retrieval with RRF, expands them using intent-aware heuristic beam search, and serializes the visited nodes into a narrative context.

\paragraph{LLM or Tool-based Retrieval.}
Some methods move beyond fixed retrievers and let an LLM or external tool actively decide where to search. This can improve recall and reasoning accuracy for difficult queries, but it is usually more expensive than direct vector or graph retrieval.
For LLM-based retrieval, MemGPT~\citep{packer2023memgpt} uses hierarchical memory without a fixed retrieval pipeline: memory tiers are exposed as tools, and the LLM selects the tier and operation under token budgets enforced by the system. MemoChat~\citep{lu2023memochat} exploits its memo structure by retrieving only the topic and summary, rather than the full topic--summary--dialogue, to reduce input length. ReadAgent~\citep{lee2024human} similarly delegates page lookup to the LLM, which decides when and which page(s) to consult.
However, while using a strong LLM can improve retrieval accuracy, it often incurs substantial overhead in both token consumption and inference latency, making it more suitable for low-frequency, high-stakes queries where correctness outweighs cost.
Besides, some methods rely on tool use for retrieval. GraphReader~\citep{li2024graphreader} predefines various tools, and employs the tools to read the memory step by step, from coarse-grained to fine-grained. D-SMART~\citep{lei2025d} lets the LLM select graph-operations such as Expand Entity and Find Path to retrieve n-hop neighbors from the global DSM and incrementally grow a task-specific subgraph, which serves as grounded context for answering.

\paragraph{Hierarchical Retrieval.}
Hierarchical retrieval follows the same coarse-to-fine principle as hierarchical memory construction. Instead of searching a large memory bank at one granularity, the agent first locates a relevant high-level segment and then drills down to finer details, reducing the search space and context inserted into the prompt. Some retrieval methods can be considered simple forms of hierarchical retrieval, such as conceptually two-layer designs. HiAgent~\citep{hu2025hiagent} can recall the trajectories by a retrieval module, when the agent needs to obtain the details of the previous subgoal. Beyond such a two-layer setup, hierarchical retrieval can be made explicit through multi-layer indexing. In H-MEM~\citep{sun2025hierarchical}, each memory embedding points to relevant sub-memories in the next layer, recursively indexing down to the last layer to retrieve relevant information, thereby accelerating retrieval. At a more system level, MemoryOS~\citep{kang2025memory} uses tier-specific retrieval: STM returns the most recent dialogue pages, MTM retrieves top-$m$ candidate segments and selects top-$k$ relevant pages within them, and LPM performs semantic search over long-term user and agent memories.
Following the same coarse-to-fine intuition, xMemory~\citep{hu2026ragagentmemoryretrieval} retrieves from high-level themes to episodes or raw messages when more detailed evidence is needed.
HyMem~\citep{zhao2026hymemhybridmemoryarchitecture} also starts from summary-level event memories and selectively inspects linked raw-dialogue evidence for more complex queries.

\paragraph{Training.}
As the memory bank grows, a fixed retriever may increasingly select memories that are semantically similar but not actually useful. Training-based retrieval addresses this by learning utility-aware selection policies, so the agent can prioritize memories that are likely to improve task success under a limited retrieval budget.
RMM \citep{tan2025prospect} adds a learnable reranker over a dense retriever and updates it online via RL using binary useful memory signals from Retrospective Reflection.
Memento \citep{zhou2025memento} learns a parametric Q-function over state–case pairs to rank and select Top-K cases, favoring historically high-reward cases over nearest neighbors.
MemRL~\citep{zhang2026memrlselfevolvingagentsruntime} introduces Two-Phase Retrieval, moving beyond passive semantic matching. It first performs similarity-based recall to obtain semantically relevant experiences, and then applies value-aware selection using learned Q-values to filter out memories that are similar but historically low-utility.

\begin{summarybox}[]{Memory Selection}
\begin{itemize}[leftmargin=0.5em]
\item \textbf{Pros}: Selects a small, task-relevant subset of memory before generation, directly reducing prompt length, attention cost, and distraction from irrelevant history.
\item \textbf{Cons}: Poor selection can be worse than no memory, because missing a critical fact may cause incorrect actions while irrelevant recalls waste context.
\item \textbf{Suitable situations}: Rule-enhanced retrieval fits clear heuristics and tight budgets; graph retrieval fits entity--relation or multi-hop evidence; LLM/tool retrieval fits low-frequency, high-stakes queries; hierarchical retrieval fits very large memory banks; training-based retrieval fits long-running systems with drifting memory distributions.
\item \textbf{Trade-offs}: Higher-capacity selectors improve recall quality but usually add latency, model calls, or training cost.
\end{itemize}
\end{summarybox}

\subsubsection{Memory Integration}
After selection, the retrieved memories still need to be integrated into the agent's reasoning context. This step affects efficiency because raw retrieved content may be verbose, redundant, or poorly aligned with the model's current decision. Efficient integration therefore filters, compresses, formats, or transforms retrieved memory so that it provides maximal guidance with minimal additional context and computation.
\paragraph{Textual Integration.} 
When memory is stored as natural language, integration mainly means deciding which small set of text to show to the backbone model and how to format it. The efficiency goal is to preserve the actionable signal of retrieved memory while minimizing prompt expansion and irrelevant context.
DC-RS~\citep{suzgun2025dynamic} integrates persistent memory by keeping a cheatsheet store, doing similarity-based retrieval, then synthesizing a compact cheatsheet that is inserted into the prompt.

Several agent-oriented systems follow the same idea but build on structured memory stores.
In Mem0~\citep{chhikara2025mem0}, each memory item is a short natural language record with metadata (time, type, source, etc.).
At inference time, the system retrieves the most relevant items and formats them as a compact memory block that is appended to the dialogue context, keeping only a handful of focused sentences in the prompt.
Taking a more structured approach, A-MEM~\citep{xu2025mem} organizes interaction history as Zettelkasten-style notes and uses a two-stage retrieval pipeline to select only a few high-utility notes; these notes are linearized into a small ``working set'' section inside the agent prompt, while the rest of the note graph remains offline.
ACE~\citep{zhang2025agentic} goes one step further and treats the agent context as an evolving playbook: it maintains a library of fine-grained strategy bullets with usage statistics, and before each episode it selects and injects only the most helpful bullets into the system instructions and memory prompts.
Similarly, for execution efficiency, agentic plan caching~\citep{zhang2025costefficientservingllmagents} caches high-level plan templates distilled from successful past executions; at serving time, a cheap keyword-based matcher looks up a matching template and a small planner LLM adapts it to the new query, replacing a fresh planning phase with a short plan-adaptation prompt.
Finally, apart from structured storage, general compression techniques are also employed to fit external information into the prompt. RECOMP~\citep{xu2023recomp} uses Retrieve–Compress–Prepend: an extractive compressor selects sentences and an abstractive compressor writes a short summary, which is prepended to the query; selective augmentation allows returning an empty string when retrieval is unhelpful.

Across these methods, textual memory integration improves efficiency by compressing long histories into task-specific snippets that fit into the prompt while retaining the main signals that drive agent behavior.

\paragraph{Latent Integration.}
Latent integration injects retrieved or stored information into the model's internal computation through compact hidden states, key--value pairs and so on. This avoids re-encoding the original text and can reduce prefill and attention costs when long-term information must be reused frequently.

One approach to latent integration is to scale latent memory capacity while keeping the GPU KV cache roughly constant.
MemoryLLM~\citep{wang2024memoryllm} inserts a trainable pool of memory tokens into every transformer layer.
During inference these tokens are processed together with the normal sequence tokens, so information stored in the memory pool can influence the hidden states at each step without extending the visible context.
Based on MemoryLLM, M+~\citep{wang2025m+} adds a CPU-resident long-term memory and a co-trained retriever that fetches a small set of relevant hidden-state memory tokens per layer during generation, enabling long-range recall with similar GPU memory overhead.

Alternatively, some latent integration methods maintain external knowledge or long context directly as compressed KV-level states, which are then integrated into the generation process via attention.
Memory$^3$~\citep{yang2024memory3} stores a KB as explicit key–value memory and, during decoding, retrieves a few entries per token block and adds their KVs to the attention KV cache, avoiding long prompts.
MemoRAG~\citep{qian2025memorag} compresses long context into a KV-cache global memory over inserted memory tokens; a lightweight memory model generates a draft answer as a retrieval clue, This design reduces the query-time long-context cost by running full long-context inference on only a few selected passages, while the rest of the corpus is accessed through compressed KV-level memory.

Compared with purely textual integration, these latent mechanisms push most long-term information into fixed-size neural states and expose them through attention, so that the cost of using long-horizon experience grows much more slowly than the length of the raw interaction history.

\subsection{Procedural Reuse via Skills}
\label{sec:skill_memory}

Beyond storing facts or interaction traces, agents can also store reusable procedures. Skills have recently emerged as a procedural form of memory for LLM agents: rather than recording what happened, they encode how an agent should act in recurring situations. This view aligns with \citet{jiang2026sokagenticskills}, who characterize agentic skills as procedural memory, and with the experience compression spectrum~\citep{zhang2026experiencecompressionspectrumunifying}, which treats episodic memories, procedural skills, and declarative rules as increasingly compressed forms of reusable agent experience.

We discuss skills in the memory section because they follow a similar efficiency lifecycle at a procedural level. Rather than recording what happened, skills encode how an agent should act in recurring situations. They are constructed from experience, selectively accessed during task execution, maintained over time, and verified before reuse. From an efficiency perspective, their value lies in amortization. A skill may require nontrivial cost to create, but once reused across similar tasks, it can reduce repeated planning, redundant tool invocation, and trial-and-error interactions. However, skills also introduce operational overhead, including retrieval, selection, updating, verification, and pruning. Thus, the key question is not simply whether skills improve task performance, but whether the downstream savings from procedural reuse can justify the cost of skill construction, access, and management.

Recent studies can be organized according to this lifecycle. 
For skill construction, several methods distill reusable skills from trajectories, interaction histories, or heterogeneous external resources. 
Trace2Skill~\citep{ni2026trace2skilldistilltrajectorylocallessons} extracts trajectory-local lessons from execution traces and consolidates them into transferable skills, making past exploration reusable for future tasks. 
Skill-Pro~\citep{mi2026skillprolearningreusableskills} learns reusable skills from experience through a non-parametric policy optimization process, turning episodic experience into executable procedural knowledge. 
SkillRL~\citep{xia2026skillrlevolvingagentsrecursive} further integrates skill abstraction into reinforcement learning, where a skill library and the agent policy are recursively improved together. 
Beyond task trajectories, AutoSkill~\citep{yang2026autoskillexperiencedrivenlifelonglearning} focuses on lifelong skill self-evolution from user interaction experience, reducing the need to repeatedly infer stable user preferences or task habits. 
SKILLFOUNDRY~\citep{shen2026skillfoundrybuildingselfevolvingagent} builds self-evolving skill libraries from heterogeneous scientific resources, converting scattered domain procedures into reusable agent skills. 
These construction methods are relevant to efficiency because they compress repeated experience or fragmented procedural knowledge into reusable skill artifacts, reducing the need to rediscover similar procedures in future tasks.

For skill access, the main efficiency challenge is how to reuse skills without overloading the context. 
As the skill library grows, naively loading all candidate skills can increase token cost, retrieval latency, and distraction from irrelevant procedures. 
SkillLens~\citep{miao2026skilllensadaptivemultigranularityskill} addresses this problem through adaptive multi-granularity skill reuse, allowing agents to reuse or adapt only the relevant parts of a skill instead of rewriting or loading the whole skill. 
Graph of Skills~\citep{liu2026graphofskillsdependencyawarestructuralretrieval} studies dependency-aware structural retrieval for massive skill libraries, retrieving bounded skill bundles under context constraints rather than exposing the agent to the full skill repository. 
These methods suggest that skill reuse is efficient only when access is selective: the agent should load enough procedural knowledge to guide action, but not so much that the skill layer itself becomes a new source of context overhead.

For skill management, recent work studies how skill libraries should evolve over time. 
EvoSkills~\citep{zhang2026coevoskillsselfevolvingagentskills} introduces co-evolutionary verification to improve self-evolving agent skills, emphasizing that automatically generated skills need validation before being reused. 
SkillClaw~\citep{ma2026skillclawletskillsevolve} considers collective skill evolution with an agentic evolver, allowing shared skill repositories to absorb recurring patterns from multiple agent interactions. 
SkillOS~\citep{ouyang2026skilloslearningskillcuration} formulates skill curation as a learnable process, where the system learns how to maintain and update a skill repository for self-evolving agents. 
From the efficiency perspective, these management mechanisms are important because an unmanaged skill library may become redundant, noisy, or outdated. 
Low-quality or irrelevant skills can increase retrieval cost, cause wrong actions, and lead to expensive error recovery.
MemSkill~\citep{zhang2026memskilllearningevolvingmemory} extends this idea from task skills to memory skills, treating memory operations such as extraction, consolidation, and pruning as reusable and evolvable procedures. 
By selecting relevant memory skills for each interaction span and refining them from difficult cases, it makes memory updates more adaptive than fixed prompt-based pipelines and amortizes memory-operation design across tasks.

Overall, skills shift part of the agent's computation from online reasoning to reusable procedural memory. 
This shift is beneficial when tasks recur, when workflows are long-horizon, or when tool-use patterns can be reused across related contexts. 
However, skill-based efficiency should be understood as an amortized trade-off. 
Constructing and maintaining skills may be costly for a single task, but the cost can be justified if the learned skills are reused across many future tasks, users, or domains. 
This also explains why recent work increasingly studies not only how to create skills, but also how to retrieve, verify, curate, and evolve them. 
For efficient agents, the central challenge is to keep the skill layer compact, reliable, and selectively accessible, so that procedural reuse reduces end-to-end cost rather than introducing another source of overhead.

\subsection{Multi-Agent Memory}
The preceding sections mainly discuss memory from a single-agent perspective, where the efficiency problem is how an agent constructs, maintains, and accesses its own experience while controlling context length, retrieval overhead, and computation cost. In multi-agent systems (MAS), memory inherits this lifecycle but changes the unit of design. The question is no longer only what an individual agent should remember, but also where reusable information should live, what should be shared across agents, what should remain local to specific roles, and how memory updates should be synchronized.

This shift introduces additional efficiency trade-offs. Naively giving every agent the full interaction trace can lead to duplicated context, repeated retrieval, and high communication cost. Conversely, relying on an overly centralized memory may introduce synchronization overhead, stale shared state, or noisy global context. Therefore, multi-agent memory must balance coordination efficiency, role specialization, consistency, communication redundancy, and cross-agent knowledge reuse. This motivates the shared, local, and mixed memory designs discussed below.

\paragraph{Shared Memory.}
Shared memory centralizes reusable information across agents, allowing the team to avoid duplicating the same observations, decisions, and intermediate results in every individual prompt. Its main efficiency benefit is reducing redundant exploration and repeated context replay, but this requires careful control of noise, staleness, and concurrent updates.
MS~\citep{gao2024memory} stores agent steps as Prompt–Answer pairs and filters them with an LLM evaluator before adding them to a shared pool, then uses accepted memories to continually refine the retriever. However, the frequent LLM-based scoring introduces substantial token and latency overhead.

To improve efficiency, recent work explores \textbf{structured shared textual memory} that supports lightweight retrieval and reduces redundant context replay.
G-Memory~\citep{zhang2025g} models multi-agent experience as a three-tier graph hierarchy of insight, query, and interaction graphs; at inference, it performs bi-directional traversal to retrieve high-level, generalizable insights together with fine-grained, condensed interaction trajectories for agent-specific working memory.
RCR-Router~\citep{liu2025rcr} maintains a Shared Memory Store of interaction history, task-relevant knowledge, and structured state representations, and performs round-wise context routing with an Importance Scorer, a Semantic Filter, and a Token Budget Allocator to minimize redundant context and token usage.
MemIndex~\citep{saleh2025memindex} adopts an intent-indexed bipartite graph architecture for memory operations in LM-based multi-agent pub/sub systems, improving storage, retrieval, update, and deletion efficiency and reporting lower elapsed time, CPU utilization, and memory usage.
Different from typical shared-memory MAS that mainly consume retrieved context, MIRIX~\citep{wang2025mirix} adopts a modular multi-agent architecture governed by a Meta Memory Manager and six Memory Managers, and uses Active Retrieval to generate a topic and inject retrieved memories into the system prompt without explicit memory-search prompts.

Beyond textual shared memory, \textbf{latent shared memory} enables agents to exchange compact internal states, reducing redundant token-level replay.
LatentMAS~\citep{zou2025latent} implements latent shared memory by having each agent perform auto-regressive latent thinking from last-layer hidden states and consolidating the resulting layer-wise KV caches into a shared latent working memory for persistent read–write sharing across agents.
KVComm~\citep{ye2025kvcomm} enables training-free online KV-cache communication by maintaining an anchor pool of shared segments and their KV offsets, then matching anchors and approximating offsets to safely reuse KV caches across new prefixes, avoiding repeated prefilling.
Cache-to-Cache~\citep{fu2026cachetocache} extends this direction to direct semantic communication between LLMs. Instead of exchanging text, it projects and fuses the source model's KV cache into the target model with a neural fuser and learnable gating, reducing token-level communication overhead.
LatentMem~\citep{fu2026latentmemcustomizinglatentmemory} customizes agent-specific latent memories in a token-efficient manner. It stores raw interaction trajectories in a lightweight experience bank and employs a memory composer to synthesize compact, role-aware latent memories, reducing information overload from fine-grained memory entries.

\begin{summarybox}[]{Shared Memory}
\begin{itemize}[leftmargin=0.5em]
\item \textbf{Pros}: Enables agents to reuse verified facts, plans, and intermediate decisions, reducing duplicated exploration and improving coordination.
\item \textbf{Cons}: Concurrent writes can introduce inconsistency, stale information, or noisy shared context, especially without provenance and access control.
\item \textbf{Suitable situations}: Best for collaborative tasks where agents need common ground, shared evidence, or reusable global state.
\item \textbf{Trade-offs}: Improves team-level efficiency but requires stronger consistency, consolidation, and permission mechanisms.
\end{itemize}
\end{summarybox}

\paragraph{Local Memory.}
Local memory keeps each agent's experience close to its own role, task, or user context. This avoids the noise and synchronization overhead of a global store, but redundancy can still accumulate within each personal memory. Therefore, local memory benefits from the same efficiency mechanisms used in single-agent settings, such as selective writing, consolidation, and capacity control.
Intrinsic Memory Agents~\citep{yuen2025intrinsic} equips each agent with a role-aligned structured memory template and updates it every turn by folding the agent’s latest output back into the same template until consensus is reached.
AgentNet~\citep{yang2025agentnet} maintains fixed-size memory modules for the router and executor, and uses dynamic memory management with signals like frequency, recency, and uniqueness to prune low-utility trajectories at capacity.
DAMCS~\citep{yang2025llm} introduces A-KGMS, consolidating experiences into a goal-oriented hierarchical knowledge graph and planning via neighborhood queries around the most recent goal node to avoid full-history sharing and reduce overhead.
\begin{summarybox}[]{Local Memory}
\begin{itemize}[leftmargin=0.5em]
\item \textbf{Pros}: Provides a lightweight, low-noise workspace tailored to each agent's role, preferences, and recent trajectory.
\item \textbf{Cons}: Useful discoveries may remain isolated, causing other agents to repeat work or make decisions without relevant evidence.
\item \textbf{Suitable situations}: Fits role-specialized agents, private user state, or subtasks where local context is more valuable than global synchronization.
\item \textbf{Trade-offs}: Reduces retrieval noise and synchronization cost, but weakens cross-agent knowledge reuse.
\end{itemize}
\end{summarybox}

\paragraph{Mixed Memory.}
Mixed memory combines shared and local stores to balance team-level reuse with agent-specific specialization. Its efficiency depends on routing: the system must decide what remains private, what should be published to shared memory, when each tier should be queried, and how redundancy across tiers should be controlled.
SRMT~\citep{sagirova2025srmt} couples each agent’s personal memory vector with a shared recurrent memory by pooling all agents’ memory vectors and letting agents cross-attend to this shared sequence, then updating their personal vectors via a memory head.
Collaborative Memory~\citep{rezazadeh2025collaborative} uses dynamic bipartite access graphs with private/shared tiers, storing fragments with immutable provenance and enforcing sharing through configurable read/write policies.
LEGOMem~\citep{han2025legomem} builds modular procedural memory with full-task memories for the orchestrator and subtask memories for task agents, comparing vanilla retrieval with Dynamic and QueryRewrite variants for finer-grained subtask memory access.

\begin{summarybox}[]{Mixed Memory}
\begin{itemize}[leftmargin=0.5em]
\item \textbf{Pros}: Combines specialized local memory with reusable shared memory, supporting both agent-specific efficiency and team-level coordination.
\item \textbf{Cons}: Requires routing decisions about what to keep local, what to publish globally, and how to resolve conflicts between memory tiers.
\item \textbf{Suitable situations}: Works well for complex multi-agent systems with both private subtask context and shared project-level knowledge.
\item \textbf{Trade-offs}: Offers the most flexible memory organization, but adds synchronization, access-control, and retrieval-routing overhead.
\end{itemize}
\end{summarybox}

\subsection{Discussion}

The above survey shows that efficient memory is not achieved by adding more storage, but by deciding when experience should be compressed, maintained, retrieved, or reused. Across construction, management, selection, and skill-based reuse, the same question recurs: which memory operation is worth its cost for future decisions?

\paragraph{Lifecycle-level budget allocation.}
Memory efficiency should be understood across the full lifecycle. Construction pays cost at write time, management pays cost between interactions, and selection pays cost at inference time. Optimizing one stage in isolation can simply move the burden elsewhere. Over-compression may require later repair, weak maintenance may leave a noisy store, and high-recall retrieval may recreate the long-context cost that memory was meant to avoid. 

\paragraph{Trade-off Between Memory Compression and Performance.} 
Although we have repeatedly emphasized that memory extraction can reduce costs such as input token usage, an unavoidable issue is that extraction may lead to the loss of critical information, which can directly degrade the agent's performance. This problem has also been noted in prior work such as AgentFold~\citep{ye2025agentfold}. LightMem~\citep{fang2025lightmem}, for instance, explicitly takes the compression rate into account. Its experimental results clearly show that excessive compression leads to poorer accuracy, whereas milder compression better preserves performance but incurs relatively higher cost. Therefore, the key question is not how to maximize compression, but how to preserve decision-relevant information under a budget. This calls for task-aware compression objectives: memories should retain causal facts, constraints, failures, and reusable procedures, rather than merely producing shorter summaries.

\paragraph{When Should Memory Be Updated?}
Memory updates are costly actions. Regarding memory management strategies, A-MEM\citep{xu2025mem} exemplifies a purely online system where memory updates occur synchronously during interaction. As demonstrated by MemoryOS\citep{kang2025memory}, such real-time updates incur frequent LLM calls per response, leading to higher latency and financial costs. By contrast, LightMem~\citep{fang2025lightmem} adopts a hybrid architecture combining a lightweight online cache with offline consolidation. This design offloads expensive computations to asynchronous offline processes, significantly reducing inference time while maintaining similar overall computational costs. This comparison highlights a fundamental trade-off: online updates ensure immediate adaptation but increase latency and cost, whereas offline updates minimize inference overhead but suffer from slower adaptation. Consequently, efficient memory systems should treat updating as a selective decision: frequent, low-risk changes can be handled by cheap rules, while expensive semantic consolidation should be triggered only when memory quality is likely to affect future decisions.

\paragraph{Evaluation Beyond Recall Accuracy.}
A deeper challenge is that current memory evaluations often emphasize downstream success or retrieval relevance, while the efficiency contribution of memory is harder to isolate. For efficient agents, a memory module should be evaluated by its marginal utility: how much additional task success it brings per token, per retrieval, per update, or per unit latency. This perspective also clarifies why memory can sometimes hurt efficiency. If a memory system retrieves many weakly relevant entries, causes the model to over-condition on stale facts, or requires frequent LLM-based maintenance, it may increase end-to-end cost despite improving recall. Future benchmarks should therefore report both effectiveness and lifecycle costs, including write cost, update frequency, retrieval latency, inserted-token budget, and failure-recovery savings.

\paragraph{Interaction with Tools and Planning.}
Memory is most valuable when it changes future behavior. In tool-using agents, memory can cache successful tool plans, API constraints, tool outputs, and common failure modes, thereby reducing redundant tool selection, parameter filling, and repeated calls. In planning agents, memory can amortize deliberation by storing reusable subgoals, skills, search outcomes, and failed branches that should be avoided. Conversely, planning and tool use also affect memory: a planner may decide when memory should be retrieved or updated, while tool observations provide new evidence that needs to be written, verified, or discarded. Therefore, the efficiency of memory should not be evaluated only by the cost of storage or retrieval. Its broader value lies in whether it reduces future planning depth, tool invocations, retries, and coordination overhead across episodes. This connection also motivates our subsequent discussion of tool use and planning, where efficiency depends not only on each component in isolation, but also on how information is reused across repeated agent execution.
\section{Efficient Tool Use}
\label{sec:tool}

\begin{figure}[t]
    \centering
    \includegraphics[width=1.0\linewidth]{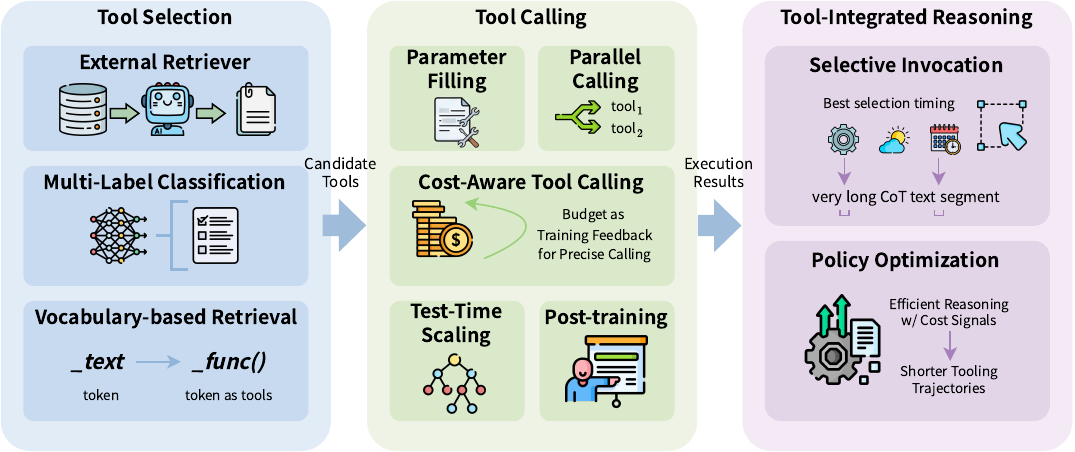}
    \caption{Efficient tool use comprises three stages: Tool Selection identifies candidate tools via retrieval or classification; Tool Calling handles parameter filling and execution with a focus on cost-aware constraints and budget feedback and Tool-Integrated Reasoning optimizes efficient reasoning trajectories through selective invocation and policy optimization.}
    \label{fig:efficient-tool-learning}
\end{figure}

\begin{table*}[t]
\centering
\caption{A summary of representative efficient tooling methods. We categorize them by tool selection, tool calling, and tool-integrated reasoning.}
\label{tab:efficient_tooling}
\resizebox{\textwidth}{!}{%
\begin{tabular}{l|l|l|l}
\toprule
\textbf{Method} & \textbf{Category} & \textbf{Core Mechanism} & \textbf{Resource Link} \\
\midrule
\multicolumn{4}{c}{\textit{Efficient Tool Selection}}\\
\midrule
ProTIP~\cite{anantha2023protipprogressivetoolretrieval} & External Retriever & Contrastive learning to correlate queries with tools & N/A \\
TinyAgent~\cite{erdogan2024tinyagent} & Multi-Label Classification & Implement a small model to select appropriate tools & \ghlink{https://github.com/SqueezeAILab/TinyAgent} \\
Tool2Vec~\cite{moon2024efficient} & Multi-Label Classification & Align tools with synthetic usage examples & \ghlink{https://github.com/SqueezeAILab/Tool2Vec} \\
ToolkenGPT~\cite{Hao2023ToolkenGPTAF} & Vocabulary-based Retrieval & Train tools as a special token & \ghlink{https://github.com/Ber666/ToolkenGPT} \\
Toolken+~\cite{nikolenko2024toolkenimprovingllm} & Vocabulary-based Retrieval & Rerank top-k tools and reject if no one is selected & N/A \\
Chain-of-Tools~\cite{wu2025cotools} & Vocabulary-based Retrieval & Leverage CoT with a huge tool pool & \ghlink{https://github.com/fairyshine/Chain-of-Tools}\\
ToolGen~\cite{wang2024toolgen} & Vocabulary-based Retrieval & Encode each tool as a separate token & \ghlink{https://github.com/Reason-Wang/ToolGen} \\
\midrule
\multicolumn{4}{c}{\textit{Efficient Tool Calling}}\\
\midrule
Toolformer~\cite{Schick2023ToolformerLM} & In-Place Parameter Filling & Leverage CoT to invoke tool calls & N/A \\
CoA~\cite{gao2025efficienttoolusechainofabstraction} & In-Place Parameter Filling & Uses symbolic abstractions for intermediate steps & N/A \\
LLMCompiler~\cite{kim2024llmcompiler} & Parallel Tool Calling & A compiler-inspired framework enabling parallel tooling & \ghlink{https://github.com/SqueezeAILab/LLMCompiler}\\
LLM-Tool Compiler~\cite{singh2024llmtoolcompilerfusedparallel} & Parallel Tool Calling & Fusing similar tools and parallel tooling & N/A \\
CATP-LLM~\cite{wu2025catp} & Parallel Tool Calling & Include cost-awareness into planning & \ghlink{https://github.com/duowuyms/OpenCATP-LLM} \\
BTP~\cite{zheng-etal-2024-budget} & Cost-Aware Tool Calling & Formulates tool calling as a knapsack problem & \ghlink{https://github.com/THUNLP-MT/BTP} \\
TROVE~\cite{wang2024troveinducingverifiableefficient} & Cost-Aware Tool Calling & Introduce compact reusable tools. & \ghlink{https://github.com/zorazrw/trove} \\
ToolCoder~\cite{ding2025toolcodersystematiccodeempoweredtool} & Cost-Aware Tool Calling & Treat tool as code generation & \ghlink{https://github.com/dhx20150812/ToolCoder} \\
ToolChain*~\cite{zhuang2023toolchainefficientactionspace} & Test-Time Scaling & Utilizes A* search to prune unproductive branches & N/A \\
OTC-PO~\cite{wang2025actingreasoningmoreteaching} & Post-training / RL & Integrates tool-use penalty into RL objective & N/A \\
ToolOrchestra~\cite{su2025toolorchestraelevatingintelligenceefficient} & Post-training / RL & Efficiency-aware rewards for specialized orchestrators & \ghlink{https://github.com/NVlabs/ToolOrchestra/} \\
\midrule
\multicolumn{4}{c}{\textit{Tool-Integrated Reasoning (TIR)}}\\
\midrule
TableMind~\cite{jiang2025tablemind} & Adaptive Search & Plan-action-reflect loop with Rank-Aware Optimization & \ghlink{https://github.com/lennendd/TableMind} \\
SMART~\cite{qian2025smart} & Boundary Awareness & CoT-based dataset to decide parametric vs. tool use & \ghlink{https://github.com/qiancheng0/Open-SMARTAgent} \\
ARTIST~\cite{singh2025agentic} & Policy Optimization & Unified agentic reasoning with outcome-based RL & N/A \\
AutoTIR~\cite{wei2025autotir} & Policy Optimization & Hybrid reward for correctness and format adherence & \ghlink{https://github.com/weiyifan1023/AutoTIR} \\
ReTool~\cite{feng2025retoolreinforcementlearningstrategic} & Code-Integrated Reasoning & Dynamic NL-code interleaving with verifiable rewards & \ghlink{https://github.com/ReTool-RL/ReTool} \\
ToolRL~\cite{qian2025toolrl} & Structured Rewards & Combines format reward with tool parameter correctness & \ghlink{https://github.com/qiancheng0/ToolRL} \\
PORTool~\cite{wu2025portool} & Step-wise Planning & Uses fork-relative advantages and decay factors & N/A \\
Agent-FLAN~\cite{chen2024agentflandesigningdatamethods} & Data Efficiency & Decomposes agent data into capability-specific subsets & \ghlink{https://github.com/InternLM/Agent-FLAN} \\
\bottomrule
\end{tabular}%
}
\end{table*}


Tool use provides an interface for LLMs to interact with the physical world and virtual environment.
In general, tools refer to search, code sandbox (interpreter), and many other general API endpoints.
To call these tools, a basic solution is to provide several candidates to the prompt, and let the LLM think and select the most suitable one with parameters filled~\cite{agent1}.
However, as the task become more complex, there would be much more tool calls.
For example, LLMs may call the search API for 600 times to resolve a deep research problem~\cite{team2025mirothinker}.
Such long trajectories extremely challenge the models' long context comprehension ability and brings enormous costs.
To this end, it is crucial to explore efficient tool use strategies.

Overall, there are two types of efficiency in tool use:
(1) Tool use itself is efficient for solving complex problems. Comparing with a task with very long CoT, tool use could efficiently optimize the length of trajectories and show the efficient reasoning process.
(2) Tool use could be optimized to call fewer tools, which reduces the cost of tool use itself. For a complex task with hundreds of tool calls, an optimal method could significantly reduces the number of tool calls. So the overall process would be even more efficient.

Figure~\ref{fig:efficient-tool-learning} presents a taxonomy of efficient tool use, which consists of three major categories: \emph{Tool Selection}, \emph{Tool Calling}, and \emph{Tool-Integrated Reasoning}. Representative methods under each category are summarized in Table~\ref{tab:efficient_tooling}. 
In a typical tool-use pipeline, candidate tools are first selected to determine when and what to invoke, and the resulting outputs are subsequently integrated into reasoning trajectories and final responses.


\subsection{Tool Selection}






For massive tool candidates from a very large pool, it is nearly impossible to stuff the prompt with thousands of tool descriptions.
To this end, it is crucial to efficiently select the most relevant tools for user queries.
We organize current tool retrieval literature into three categories:
(1) \textbf{External Retriever}: a independent retriever model which embeds user queries and tool descriptions and calculates the affinity scores (e.g. cosine similarity) to select top-$k$ relevant tools as candidates;
(2) \textbf{Multi-Label Classification}: for a fixed size of tool sets, the tool selection process could be formulated as a multi-label classification problem, which directly predicts relevant tools;
and (3) \textbf{Vocabulary-based Retrieval}: tools are embedded as special tokens into the model's vocabulary, and the model would enter a tool call mode when generating such tool tokens.
We introduce the above three categories of tool selection strategies in this section as below.

\paragraph{External Retriever.}
Instead of including the entire tool set, many approaches rely on an external retriever for tool selection. 
External tool retrieval can be improved from retriever-side advances that redesign the retrieval pipeline or strengthen retrievers and rerankers, and from tool-side enhancements that refine tool descriptions and documentation to make the retrieval corpus easier to match, boosting both accuracy and efficiency.

On the retriever side, ProTIP~\cite{anantha2023protipprogressivetoolretrieval} utilizes a contrastive learning-based method to embed user queries and tool descriptions into the semantic space.
After a tool is selected, ProTIP subtracts the query embedding by the selected tool's representation and selects tools on other subtasks.
Such a progressive design makes ProTIP efficient to avoid explicit task decomposition overhead.
In AnyTool~\citep{pmlr-v235-du24h}, retrieval is organized hierarchically and inspired by a divide-and-conquer strategy, narrowing the search space and thereby improving retrieval efficiency.

On the tool side, DRAFT~\citep{quexploration} refines tool documents via self-driven interactions to improve external tool retrieval, while boosting efficiency by reducing token overhead and stopping refinement at convergence.

In addition, some recent systems combine both directions. Toolshed~\citep{lumer2024toolshed} stores enriched tool representations in a tool knowledge base and uses RAG-tool fusion before, during, and after retrieval to scale external tool selection, while controlling top-k to curb token growth and improve efficiency.
Similarly, ToolScope~\citep{liu2025toolscopeenhancingllmagent} uses ToolScopeMerger with Auto-Correction to compress tool descriptions and reduce input tokens, and ToolScopeRetriever to hybrid-retrieve top-k tools that fit the LLM context window, improving tool-use quality while boosting efficiency and scalability.

\paragraph{Multi-Label Classification (MLC).}
Instead of ranking-based retrieval, MLC-based methods treat tool selection as a classification task. TinyAgent~\cite{erdogan2024tinyagent} is designed to conduct tool calling on edge devices which pursue extreme efficiency, and it formulates the tool selection task as a multi-label classification problem.
For a user query, TinyAgent applies the DeBERTa-v3 small model as the encoder and output the probability distribution for all available tools.
Tools with a probability higher than 50\% are recognized as relevant ones and will be selected accordingly.
Since only a small fraction of tool descriptions are put into the prompt, it efficiently reduces nearly half of the prompt size.
Similar to TinyAgent, \citet{moon2024efficient} find MLC-based tool retrieval efficient, but such a task formulation could not handle the growing number of tools, and any updates would require a model re-training.
Therefore, they propose Tool2Vec, a two-stage retrieval with a reranker for analyzing fine-grained tool-query interactions.
To fill in the semantic gap where natural user queries may not directly align with tool descriptions, the authors generate tool embeddings based on synthetic usage examples rather than static description.

\paragraph{Vocabulary-based Retrieval.}
Besides direct retrieving from candidates by external retriever and MLC, tool selection could also be formulated as a token prediction task, where tools are stored in the vocabulary as special tokens.

ToolkenGPT~\cite{Hao2023ToolkenGPTAF} regards massive external tools as learnable token embeddings (aka. ``toolken''), so that the target tool could be selected as a normal next token prediction process.
Compared with Toolformer~\cite{Schick2023ToolformerLM} that selects tools by predicting a whole trajectory with special characters, this approach is highly efficient since it only trains the added tool embeddings and keeps other model parameters frozen.
Furthermore, it bypasses the window constraint of in-context tool selection and retains a shorter prompt.
Building on this foundation, Toolken+~\cite{nikolenko2024toolkenimprovingllm} enhances ToolkenGPT by introducing an extra reranking step and a rejection toolken, which improves the overall performance and reduces the hallucination rate.
Toolken+ also demonstrates a tradeoff between efficiency and efficacy, which could be simply tuned from the number of reranking candidates.
Although ``toolkens'' are efficient for massive tool selection, it requires constructing data samples for supervised fine-tuning and suffers from generalization problems on unseen tools.
Similarly, ToolGen~\citep{wang2024toolgen} assigns each tool a unique tool token and trains the model to turn tool retrieval and calling into a unified generation task. By representing a tool with a single token, it is claimed to shorten generation and potentially reduce inference overhead but may be costly at training phase.
From a different perspective on efficiency, \citet{xu2024concise} proposes selective compression and block compression for tool use: they preserve key information (e.g., tool and parameter names) as raw text while compressing the remaining documentation into fixed-length soft tokens per block. The soft tokens can be precomputed and cached offline, reducing prompt length and improving token efficiency at inference.
To tackle the generalization problem, CoTools~\cite{wu2025cotools} shrinks the number of toolkens to only one and applies a retriever to calculate the similarities between current toolken's representation and all the candidates.

From these literature, we find vocabulary-based methods are an efficient option for tool selection.
However, it may suffer from inaccurate invocation timing and poor generalization to unseen new tools, which is less functional for extensive tool updating scenarios.

\begin{summarybox}[]{Tool Selection}
\begin{itemize}[leftmargin=0.5em]
\item \textbf{Pros}: Narrows a large tool pool before generation, reducing prompt length, decoding ambiguity, and unnecessary tool exploration; retriever-based selection can also generalize to newly added tools.
\item \textbf{Cons}: External retrievers may introduce their own inference cost, while MLC- or vocabulary-based selection usually requires fine-tuning and can be less flexible when tools change.
\item \textbf{Suitable situations}: External retrievers fit dynamic or very large tool collections; MLC and vocab-based tool retrieval fit relatively fixed tool sets where low-latency selection is more important.
\item \textbf{Trade-offs}:  Broad selection preserves tool coverage but increases prompt and planning cost, while aggressive selection improves efficiency but may filter out necessary tools. Tiered selection can balance low-latency access to frequent tools with retrieval flexibility for rare or new tools.
\end{itemize}
\end{summarybox}

\subsection{Tool Calling}

Once candidates are selected, the efficiency of the invocation process becomes critical for real-time agentic interactions.
Here, efficiency is better understood at the trajectory level rather than by the cost of an isolated call, since planning, verification, and orchestration may reduce failed calls, retries, and unnecessary environment interactions.
\paragraph{In-Place Parameter Filling.}
In-place tool calling is a paradigm where the model directly fills the tool's parameters during the response generation process.
Toolformer~\cite{Schick2023ToolformerLM} incorporates tool calling within the CoT path, and fills parameters during the response generation process.
It is efficient to obtain the final results once the closure of the tool call is reached.
\citet{gao2025efficienttoolusechainofabstraction} proposes CoA, which shares the similar idea but reduces the response time by providing more accurate tool call results.
Instead of directly calculating the final results, CoA introduces symbolic abstractions to represent the intermediate steps, which are later substituted with the actual results during the response generation process.
From the experimental results, CoA performs better while reducing more than 30\% inference time than Toolformer.

\paragraph{Parallel Tool Calling.}
For a complex tasks that incorporates multiple tools, traditional sequential style calling may hurt efficiency since LLMs have to wait for the latest tool call's response.
However, there are multiple tasks that could be done in parallel~\cite{zhang2025paralleltaskplanning}.
For example, when collecting weather information for all cities in a province, an agent does not need to invoke a weather-query tool sequentially for each city.
Instead, these independent tool calls can be executed in parallel, which significantly reduces the overall task-solving time.
LLMCompiler~\cite{kim2024llmcompiler} introduces a compiler-inspired framework that formulates execution plans, dispatches tasks, and executes functions in parallel.
This achieves improvements in latency, cost, and overall accuracy against the traditional sequential tool execution approach.
Building on this parallelization paradigm, LLM-Tool Compiler~\cite{singh2024llmtoolcompilerfusedparallel} further optimizes efficiency by selectively fusing similar tool operations at runtime, which increases parallel tool calls while reducing token consumption and latency.
W\&D~\citep{lin2026wdscalingparalleltoolcalling} extends this idea to deep research agents by scaling width through parallel tool calling, allowing multiple tools to be invoked within a single reasoning step to improve information-gathering efficiency and reduce interaction turns.
Complementing with the above methods, CATP-LLM~\cite{wu2025catp} addresses the execution cost by incorporating cost-awareness into the planning process.
It designs a multi-branch planning language and employs cost-aware offline reinforcement learning to fine-tune models, enabling high-quality generation with economic constraints.

\paragraph{Cost-Aware Tool Calling.}
Like we have introduced about CATP-LLM in the above paragraph, cost could be a special reward for training efficient tool calling models.
Budget-Constrained Tool Use with Planning (BTP)~\cite{zheng-etal-2024-budget} first formulates tool calling as a knapsack problem, which utilizes dynamic programming to pre-compute how often each tool would be invoked under a hard budget, thereby turning cost control into a forward-looking plan.
Building on this planning strategy, \citet{xu-etal-2025-alignment} estimates LLM confidence via consistency-based sampling strategy to let the model trigger a tool under a certainty-cost optimal condition.
This method could reduces the number of tool calls, thereby boosting the overall improvements.
From a broader system perspective, \citet{wu-etal-2025-joint} reduces redundant calls by jointly updating prompt strategy and tool documentations.
It complements the above cost-aware planning and confidence-based gating with context-level efficiency.

Beyond directly constraining invocation budgets, recent research also explores improving efficiency through alternative paradigms, such as function induction, code generation, and model distillation. TROVE~\cite{wang2024troveinducingverifiableefficient} introduces a training-free paradigm that incrementally builds and trims a compact toolbox of reusable functions, showing that online induction can improves the accuracy without extra training data.
ToolCoder~\cite{ding2025toolcodersystematiccodeempoweredtool} extends this idea by formulating tool use as an end-to-end code generation task, which converts tasks in natural languages into Python code.
This method boosts the success rates while keeping small API usage cost.
Focusing on the deployment cost, \citet{kang2025distillingllmagentsmall} proposes to distill LLM's knowledge into small language models with retrieval and code interpreter tools, which enables small models competitive with larger ones.

\paragraph{Efficient Test-Time Scaling.}
For effective tool calling, a viable solution is tree search-based strategies, where the model may plan a tree of tool calls and select the most promising path~\cite{wu2024toolplanner}.  
However, such methods are computationally expensive since they may need trail-and-error to explore the entire tree.
Instead of extensive tree traversal, ToolChain*~\cite{zhuang2023toolchainefficientactionspace} utilizes the A* search strategy to efficiently navigate complex action spaces.
This method boosts the efficiency by employing task-specific cost functions to prune wrong branches earlier and only requires single-step node expansions.

Recent work further optimizes tool-use trajectories through explicit orchestration and search. Utility-Guided Agent Orchestration~\citep{liu2026utilityguidedagentorchestrationefficient} formulates tool use as a quality-cost decision problem, selecting among responding, retrieving, invoking tools, verifying intermediate results, and stopping according to estimated gain, step cost, uncertainty, and redundancy. ToolTree~\citep{yang2026tooltreeefficientllmagent} instead casts multi-step tool use as an MCTS-inspired search over executable tool-use trajectories, using dual-stage LLM feedback and bidirectional pruning to allocate exploration to promising tool paths. These methods show that efficient tool calling requires controlling not only which tool to call, but also when to stop, verify, or search over alternative tool-use trajectories.

\paragraph{Efficient Tool Calling with Post-training.}
To mitigate the latency, computational overhead, and training cost associated with multi-step tool interactions, recent research has increasingly focused on optimizing tool-calling efficiency through post-training.
Specifically, reinforcement learning has emerged as a primary mechanism for teaching models to strategically balance task success with resource parsimony.
OTC-PO~\cite{wang2025actingreasoningmoreteaching} promotes action-level efficiency by integrating a tool-use penalty into the reinforcement learning objective, effectively training models to minimize redundant tool calls without sacrificing answer correctness.
Building on the optimization of agentic workflows, ToolOrchestra~\cite{su2025toolorchestraelevatingintelligenceefficient} leverages efficiency-aware rewards within an RL framework to train specialized orchestrators that achieve superior task performance at a fraction of the computational cost of general-purpose large language models.
Beyond optimizing the learned tool-use policy, \citet{liu2026effectivenessefficiencyagentictoolcalling} improve the efficiency of RL-based tool-calling training itself by filtering low-signal prompts and down-sampling less informative rollouts, reducing training cost while preserving tool-calling performance.
Complementing these optimization-oriented approaches, ToolRM~\cite{agarwal2025toolrmoutcomerewardmodels} uses outcome-based reward models to support data-efficient fine-tuning and inference-time scaling, encouraging models to favor effective and concise tool-calling trajectories.

\begin{summarybox}[]{Tool Calling}
\begin{itemize}[leftmargin=0.5em]
\item \textbf{Pros}: Improves trajectory-level efficiency by making tool use more selective, coordinated, and cost-aware, reducing unnecessary calls, redundant interactions, and avoidable retries.
\item \textbf{Cons}: Some methods introduce additional reasoning, sampling, orchestration, or verification overhead; parallel and search-based calling may also become inefficient when dependency analysis or task decomposition is inaccurate.
\item \textbf{Suitable situations}: Most useful for agents that repeatedly interact with external tools, especially when tool calls are expensive, latency-sensitive, or dependent on previous observations.
\item \textbf{Trade-offs}: More control over tool calling can reduce wasted actions and improve success, but excessive control may shift cost from tool execution to planning, routing, or verification.
\end{itemize}
\end{summarybox}


\subsection{Tool-Integrated Reasoning}







The emergence of agents marks a crucial shift from reliance on static internal knowledge toward adaptive, multi-turn reasoning, which is necessary for achieving both high accuracy and computational efficiency in complex problem-solving \cite{ma2024sciagent,ruan2023tptu,qu2025survey}. 
Traditional, rigid programmatic workflows or purely text-based methods often fail on tasks requiring numerical precision or dynamic adaptation, thereby constraining the development of truly autonomous reasoning capabilities. 

\paragraph{Selective Invocation.} 
The quest for efficient agents begins with establishing a robust capability to invoke tools only when strictly necessary, thereby minimizing redundant computations. Traditional rigid workflows often lead to excessive interactions. 
The TableMind framework \cite{jiang2025tablemind} addresses this by presenting an autonomous programmatic agent specifically tailored for tool-augmented table reasoning. Architecturally, TableMind utilizes an iterative plan-action-reflect loop, where the agent first decomposes a problem, then generates and executes precise code within a secure sandbox environment. 
TableMind employs a two-stage training paradigm: Supervised Fine-Tuning (SFT) serves as a vital warm-up phase to establish foundational tool usage patterns and master the necessary syntax for the iterative cycle, thereby mitigating the instability associated with starting subsequent Reinforcement Learning from a cold policy. 
To further refine the efficiency of tool invocation, \citet{qian2025smart} first constructs a dataset called SMART with CoT detailing the necessity of each tool call, and they use the dataset to fine-tune a model that efficiently decides whether to use their parametric knowledge or external tools. 
Agent-FLAN~\citep{chen2024agentflandesigningdatamethods} separates format-following agent data from general reasoning data and further decomposes agent data into capability-specific subsets, which improves performance with fewer training tokens.
Tool-Star~\citep{dong2025toolstarempoweringllmbrainedmultitool} further improves tool-use selectivity through curated tool-integrated reasoning data. Its synthesis and filtering pipeline reduces low-quality or redundant tool-use demonstrations, encouraging agents to invoke tools only when they are likely to be useful.

\paragraph{Cost-Aware Policy Optimization.}
After supervised warm-up, Reinforcement Learning (RL) becomes important for optimizing complex multi-step tool-use policies, especially when the agent must maintain reasoning quality, valid tool formats, and efficient execution trajectories.
To prioritize high-quality trajectories, TableMind~\citep{jiang2025tablemind} employs Rank-Aware Policy Optimization (RAPO), which identifies misaligned trajectories and applies rank-aware advantage weighting to guide the model toward consistent answers.
ARTIST~\citep{singh2025agentic} tightly couples agentic reasoning with outcome-based RL, enabling models to learn tool-use strategies without restrictive step-level supervision.
Similarly, ReTool~\citep{feng2025retoolreinforcementlearningstrategic} integrates a code interpreter into the reasoning loop, allowing the model to interleave natural language reasoning with executable code and discover strategies through verifiable rewards.
ToolRL~\citep{qian2025toolrl} further introduces structured rewards for tool calls by combining format validity, correctness, and parameter matching, improving the reliability of each tool invocation.

A central efficiency objective in this line of work is to reduce unnecessary tool use and shorten tool-use trajectories.
Methods such as A$^2$FM~\citep{chen2025a2fmadaptiveagentfoundation} and IKEA~\citep{huang2025reinforced} balance internal knowledge with external retrieval.
A$^2$FM uses Adaptive Policy Optimization (APO) with a self-adaptive router to decide whether to answer directly or invoke tools, while IKEA trains an adaptive search agent to rely on internal knowledge first and call search APIs only when necessary.
Other methods encode tool-use efficiency more explicitly into reward design or trajectory construction.
AutoTIR~\citep{wei2025autotir} discourages unnecessary tool usage through reward penalties, and OTC-PO~\citep{wang2025actingreasoningmoreteaching} encourages trajectories that produce correct answers with fewer tool calls.
SWiRL~\citep{goldie2025syntheticdatageneration} filters redundant actions during parallel trajectory generation, while PORTool~\citep{wu2025portool} assigns step-wise importance with a decay factor $\gamma$, giving higher weight to decisions closer to the final outcome and favoring solutions with fewer tool-call steps.

Recent methods further refine this trajectory-level optimization by localizing which part of the tool-use process should be updated.
EvoTool~\citep{yang2026evotoolselfevolvingtoolusepolicy} decomposes the tool-use policy into Planner, Selector, Caller, and Synthesizer modules, and uses blame-aware mutation to update the module most responsible for a failed trajectory.
ELPO~\citep{liang2026learningirrecoverableerrorlocalizedpolicy} localizes the first irrecoverable error with binary-search rollout trees and turns it into step-level learning signals.
These methods complement earlier reward- and trajectory-level optimization by targeting the specific module or step that causes inefficient or unsuccessful tool use, rather than treating the whole trajectory as a monolithic success or failure.

\begin{summarybox}[]{Tool-Integrated Reasoning}
\begin{itemize}[leftmargin=0.5em]
\item \textbf{Pros}: Improves tool-augmented reasoning by teaching agents when to invoke tools, how to produce valid tool calls, and how to avoid redundant or ineffective tool-use trajectories.
\item \textbf{Cons}: Requires high-quality tool-use data, reliable feedback signals, reward design, or rollout-based optimization. Poorly designed objectives may lead to tool overuse, tool underuse, format overfitting, or shorter but less reliable trajectories.
\item \textbf{Suitable situations}: Best for tasks where the agent must learn a nontrivial tool-use policy, including when to use tools, how to format calls, how to interpret observations, and how to balance task success with tool-call cost.
\item \textbf{Trade-offs}: Tool-integrated optimization can improve selectivity, correctness, and trajectory efficiency, but shifts cost to data synthesis, sandboxed execution, RL rollouts, and fine-grained credit or blame assignment.
\end{itemize}
\end{summarybox}

\subsection{Discussion}

Efficient tool use is not equivalent to using fewer tools in all cases. Instead, the central problem is to decide which external actions are worth their cost. Tool selection reduces the action space before generation, tool calling optimizes the execution trajectory, and tool-integrated reasoning decides when external evidence or computation should enter the reasoning loop. Together, these directions show a shift from merely enabling tool use to optimizing the full interaction loop under latency, token, and monetary budgets.

\paragraph{From Tool Availability to Tool Utility.}
A large tool library can improve coverage but also increases selection difficulty, prompt length, and the risk of invoking irrelevant tools. Thus, the efficiency bottleneck is not only the cost of a single API call, but also the cost of representing, searching, and deciding among tools. Retrieval-based selectors preserve flexibility for dynamic tool sets, while MLC- or vocabulary-based approaches reduce latency for stable tool inventories. This suggests a useful design principle: frequent and high-utility tools can be internalized into faster specialized selectors, whereas long-tail or volatile tools should remain accessible through an elastic retrieval layer.

\paragraph{Knowing When Not to Call Tools.}
A deeper issue is calibration. Tool-integrated agents must learn the boundary between what the model can reliably answer internally and what requires external execution or verification. Over-triggering tools wastes latency and money, while under-triggering tools risks hallucination or invalid actions. Cost-aware tool training and selective invocation methods therefore point toward a broader objective: optimize the accuracy-per-cost frontier rather than raw task accuracy. In this view, an efficient tool user is not the agent that calls the fewest tools, but the agent that calls tools only when the expected reduction in uncertainty or error justifies the cost.

\paragraph{Parallelism and Dependency Awareness.}
Parallel tool calling reduces wall-clock latency, but it is not automatically efficient. When subtasks are independent, parallelism can turn sequential waiting time into concurrent execution. When dependencies are misidentified, however, parallel branches may generate inconsistent intermediate results that require additional reconciliation. Therefore, parallel tool use should be paired with dependency-aware planning: the agent must distinguish actions that can be safely executed in parallel from those that require sequential evidence accumulation.

\paragraph{Interaction with Memory and Planning.}
Tool efficiency is closely tied to memory and planning. Memory can store successful tool-use templates, API constraints, parameter choices, and failure cases, reducing future selection and calling cost. Planning determines whether a task should be solved through direct reasoning, tool execution, or a decomposed multi-step workflow. Conversely, tools can externalize parts of reasoning, allowing the planner to rely on calculators, search engines, code interpreters, or databases rather than expanding the internal chain of thought. Tool observations also become memory candidates: useful outputs may be cached for reuse, while failed calls can inform future avoidance. Therefore, a method categorized as efficient tool use may obtain its end-to-end efficiency by reducing planning search or improving memory quality. This interaction suggests that future tool-using agents should jointly optimize tool choice, memory reuse, and planning depth, rather than treating tool calls as isolated actions.

\section{Efficient Planning}
\label{sec:efficient_planning}

Efficient planning studies how agents allocate limited reasoning resources during task solving. 
Unlike classical planning, which often abstracts away the cost of search or optimization, LLM-based agents expose planning costs through token usage, latency, tool calls, environment interactions, and communication overhead.
Thus, planning is not only about generating a valid action sequence, but also about deciding how much reasoning, search, verification, and coordination should be spent before acting.

\begin{figure}[t]
    \centering
    \includegraphics[width=1\linewidth]{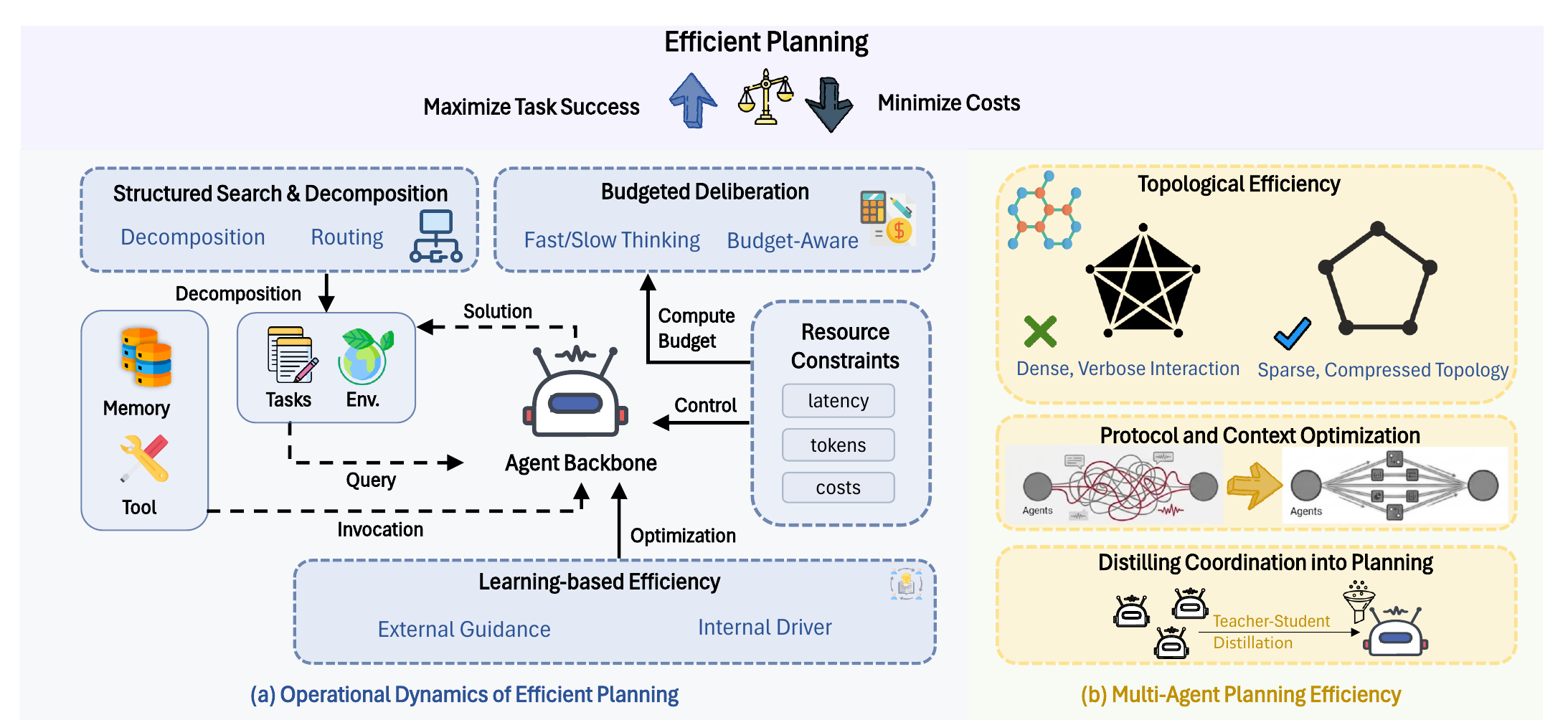}
    \caption{\textbf{Overview of Efficient Planning.} It aims to maximize task success while minimizing costs. \textbf{(a)} Single-agent methods optimize inference strategies (control, search, decomposition) or evolve via learning (policy, memory). \textbf{(b)} Multi-agent methods reduce overhead via topological optimization, context optimization, and coordination distillation.}
    \label{fig:planning_fig}
\end{figure}

From this perspective, planning becomes a form of online compute allocation. Deeper reasoning or broader collaboration may improve task success, but it can also introduce extra search, longer contexts, redundant communication, and delayed execution. Efficient planning therefore aims to balance the benefit of further deliberation against its computational and interaction cost. As illustrated in Figure~\ref{fig:planning_fig}, efficient planning can be broadly divided into \textbf{Single-Agent Planning}, which optimizes individual deliberation through inference-time strategies and learning-based evolution, and \textbf{Multi-Agent Collaborative Planning}, which reduces coordination overhead in distributed systems. Representative methods in each category are summarized in Table~\ref{tab:efficient_planning_summary}. In the remainder of this section, we review efficient planning from these two perspectives.

\begin{table*}[t]
\centering
\caption{A summary of representative efficient planning methods. We categorize Single-Agent methods into \textbf{Inference-Time Strategy} (Adaptive Control, Search, Decomposition) and \textbf{Learning-based Evolution} (Policy, Memory), alongside \textbf{Multi-Agent Collaborative Efficiency}.}
\label{tab:efficient_planning_summary}
\resizebox{\textwidth}{!}{%
\begin{tabular}{l|l|l|l}
\toprule
\textbf{Method} & \textbf{Category} & \textbf{Core Mechanism} & \textbf{Resource Link} \\
\midrule
\multicolumn{4}{c}{\textbf{\textit{Single-Agent: Inference-Time Strategy (Search \& Control)}}}\\
\midrule
SwiftSage~\cite{lin2023swiftsagegenerativeagentfast} & Adaptive Control & Fast/Slow Dual-process (System 1 + 2) & \ghlink{https://github.com/SwiftSage/SwiftSage}\\
Budget-Aware~\cite{liu2025budgetawaretooluseenableseffective} & Adaptive Control & Budget-constrained tool policy allocation & N/A \\
Reflexion~\cite{shinn2023reflexionlanguageagentsverbal} & Adaptive Control & Verbal reinforcement from prior failures & \ghlink{https://github.com/noahshinn/reflexion} \\
LATS~\cite{zhou2024languageagenttreesearch} & Tree Search & MCTS with self-reflection & \ghlink{https://github.com/lapisrocks/LanguageAgentTreeSearch} \\
ToolChain*~\cite{zhuang2023toolchainefficientactionspace} & Tree Search & A* search with learned cost pruning & N/A \\
CATS~\cite{zhang2025costaugmented} & Tree Search & Cost-aware pruning in tree search & N/A \\
ReWOO~\cite{xu2023rewoo} & Decomposition & Planner-Worker-Solver separation & \ghlink{https://github.com/billxbf/ReWOO} \\
HuggingGPT~\cite{shen2023hugginggptsolvingaitasks} & Decomposition & Routing tasks to specialized models & \ghlink{https://github.com/microsoft/JARVIS} \\
Alita~\cite{qiu2025alita} & Decomposition & MCP brainstorming \& subtasking & \ghlink{https://github.com/CharlesQ9/Alita} \\
\midrule
\multicolumn{4}{c}{\textbf{\textit{Single-Agent: Learning-based Evolution (Policy \& Memory)}}}\\
\midrule
QLASS~\cite{lin2025qlass} & Policy Optimization & Q-Value critic for search guidance & N/A \\
ETO~\cite{song2024trialanderroreto} & Policy Optimization & Trial-and-error preference learning (DPO) & \ghlink{https://github.com/Rafa-zy/QLASS} \\
VOYAGER~\cite{wang2023voyageropenendedembodiedagent} & Memory \& Skill & Iterative skill library construction & \ghlink{https://github.com/MineDojo/Voyager} \\
GAP~\cite{wu2025gap} & Memory \& Skill & Graph-based decomposition \& parallelism & \ghlink{https://github.com/WJQ7777/Graph-Agent-Planning} \\
RLTR~\cite{li2025encouraginggoodprocessesneed} & Policy Optimization & Process-level reward training & N/A \\
Planning w/o Search~\cite{hong2025planningsearchrefiningfrontier} & Policy Optimization & Offline goal-conditioned critic & \weblink{https://jxihong.github.io/pnlc_website/} \\
\midrule
\multicolumn{4}{c}{\textbf{\textit{Multi-Agent: Collaborative Efficiency}}}\\
\midrule
Chain-of-Agents~\cite{zhang2024chain} & Topology & Sequential context passing (Linear complexity) & \ghlink{https://github.com/AdamCodd/Chain-of-agents} \\
MacNet~\cite{qian2024scaling} & Topology & DAG-based topological ordering & \ghlink{https://github.com/OpenBMB/ChatDev/tree/macnet} \\
AgentPrune~\cite{zhang2024cut} & Topology & Learned pruning of communication edges & \ghlink{https://github.com/yanweiyue/AgentPrune} \\
MARS~\cite{wang2025mars} & Topology & Reviewer-Meta-Reviewer pipeline (No debate) & \ghlink{https://github.com/xwang97/MARS} \\
CodeAgents~\cite{yang2025codeagents} & Protocol & Structured pseudocode interaction & \ghlink{https://anonymous.4open.science/r/CodifyingAgent-5A86} \\
Free-MAD~\cite{cui2025free} & Protocol & Prompt-optimized critical reasoning & N/A \\
MAGDI~\cite{chen2023magdi} & Distillation & Distilling interaction graphs into student & \ghlink{https://github.com/dinobby/MAGDi} \\
D\&R~\cite{zhou2025debate} & Distillation & Distilling debate traces via DPO & N/A \\
\bottomrule
\end{tabular}%
}
\end{table*}
\subsection{Single-Agent Planning Efficiency}
\label{subsec:single_agent_planning}


Single-agent efficiency focuses on minimizing the computational cost, measured in tokens, latency, or search steps, required to reach a valid solution. We categorize these methods into \textit{inference-time strategies}, which optimize the planning process on-the-fly, and \textit{learning-based evolution}, which improves the agent's intrinsic planning capabilities.

For single agents, efficient planning is less about always producing shorter reasoning traces and more about deciding when additional deliberation is worth paying for. Some tasks benefit from deeper search because it avoids costly retries, while others are better handled by fast heuristics or cached skills. This makes adaptive compute allocation a central design principle across the methods below.

\paragraph{Inference Strategy I: Adaptive Budgeting and Control.}
A key strategy is \textit{selective deliberation}, allocating computational effort non-uniformly. Architectures like SwiftSage~\cite{lin2023swiftsagegenerativeagentfast} separate fast behaviors from slower planning, defaulting to heuristics unless structured reasoning is required. This can be framed as learning when to invoke a costly planner versus a reactive policy~\cite{paglieri2025learningplanefficientlyallocating}, or dynamically adjusting tool strategies based on budget constraints~\cite{liu2025budgetawaretooluseenableseffective}. 

Recent work further makes this control more fine-grained by treating reasoning depth as a step-level decision. Ares~\citep{yang2026aresadaptivereasoningeffort} trains a lightweight router to select the lowest sufficient reasoning effort for each agent step, while Think Fast and Slow~\citep{yang2026thinkfastslowsteplevel} learns to switch among multiple cognitive depths, ranging from instinctive responses to strategic planning. These methods extend fast/slow agent designs by reserving expensive reasoning for difficult decisions rather than applying deep deliberation uniformly across the whole trajectory.

Efficiency is also gained by preventing redundant failures; methods like Reflexion~\cite{shinn2023reflexionlanguageagentsverbal} and ReST~\cite{aksitov2023restmeetsreactselfimprovement} use verbal reinforcement or iterative refinement to amortize failure analysis, lowering cumulative interaction costs.

\paragraph{Inference Strategy II: Structured Search.}
The combinatorial explosion of action spaces presents a significant bottleneck. To address this, methods adapt formal search algorithms to prune feasible trajectories. Language Agent Tree Search (LATS)~\cite{zhou2024languageagenttreesearch} reframes agent rollouts as Monte Carlo Tree Search, enabling self-reflection to guide exploration. Building on this, CATS~\cite{zhang2025costaugmented} integrates cost-awareness directly into the search tree, pruning expensive branches early. In tool-rich environments, ToolChain*~\cite{zhuang2023toolchainefficientactionspace} applies A* search to navigate the action space, while retrieval-based approaches like ProTIP~\cite{anantha2023protipprogressivetoolretrieval} reduce decision complexity by only surfacing relevant tools during the planning phase.

\paragraph{Inference Strategy III: Task Decomposition.}
Explicitly breaking down complex tasks can reduce redundant reasoning over long trajectories. 
ReWOO~\cite{xu2023rewoo} and Alita~\cite{qiu2025alita} decouple planning from execution by generating intermediate blueprints, while Task-Decoupled Planning~\citep{li2026entangledplanningtaskdecoupledplanning} further decomposes long-horizon tasks into DAG-structured sub-goals and restricts planning or replanning to scoped subtask contexts. 
These designs reduce token overhead and recovery cost by avoiding repeated reasoning over a monolithic trajectory.

Task decomposition also facilitates routing. 
HuggingGPT~\cite{shen2023hugginggptsolvingaitasks} and ReSo~\cite{zhou2025resorewarddrivenselforganizingllmbased} dispatch sub-tasks to specialized models, while BudgetMLAgent~\cite{gandhi2025budgetmlagentcosteffectivellmmultiagent} optimizes agent routing under cost considerations. 
In embodied settings, AutoGPT+P~\cite{birr2024autogptpaffordancebasedtaskplanning} grounds decomposition in environmental affordances to improve feasibility.

\paragraph{Learning-Based Evolution: Policy Optimization.}
Agents can learn to internalize planning logic. This is driven by external critics, such as QLASS~\cite{lin2025qlass} or offline value functions~\cite{hong2025planningsearchrefiningfrontier}, that guide the planner toward high-value actions. Alternatively, learning acts as an \textit{internal driver}: ETO~\cite{song2024trialanderroreto} refines policies via trial-and-error preference learning (DPO). To improve sample efficiency, methods like RLTR~\cite{li2025encouraginggoodprocessesneed} and Planner-R1~\cite{zhu2025plannerr1rewardshapingenables} utilize process-level rewards, providing feedback on the reasoning sequence rather than just the final outcome.
In long-horizon web reasoning, WebAnchor~\citep{yu2026webanchoranchoringagentplanning} further emphasizes the importance of early planning decisions by optimizing the first-step plan with dense rubric-based rewards and then aligning execution with the anchored plan using sparse rewards.
Together, these methods reduce inefficient exploration by making planning policies more sensitive to intermediate progress, early decision quality, and downstream execution cost.
\paragraph{Learning-Based Evolution: Memory and Skill Acquisition.}
Efficiency can be amortized by externalizing successful plans. VOYAGER~\cite{wang2023voyageropenendedembodiedagent} builds a library of reusable skills to avoid re-planning. Graph-based representations also support this: GraphReader~\cite{li2024graphreader} and other graph-enhanced models~\cite{lin2024graphenhancedlargelanguagemodels} leverage structured memory for long-context retrieval, while GAP~\cite{wu2025gap} identifies parallelizable actions. Ultimately, frameworks like Sibyl~\cite{wang2024sibylsimpleeffectiveagent} demonstrate that efficiency is an emergent property, where improved memory structure directly reduces the cognitive load of future planning.

\begin{summarybox}[]{Single-Agent Strategies}
\begin{itemize}[leftmargin=0.5em]
\item \textbf{Pros}: Adaptive control reduces unnecessary deliberation, structured search improves exploration, decomposition limits repeated context processing, and learning-based evolution amortizes planning cost over future tasks.
\item \textbf{Cons}: Control policies can misfire, search adds branching overhead, decomposition may propagate early errors, and learning or memory components require training and maintenance.
\item \textbf{Suitable situations}: Best when one agent must solve long-horizon tasks under a limited compute or interaction budget, especially when the task can be decomposed or guided by reusable experience.
\item \textbf{Trade-offs}: More planning depth usually improves robustness, but the marginal benefit decreases once search, reflection, or decomposition costs exceed the value of the refined plan.
\end{itemize}
\end{summarybox}

\subsection{Multi-Agent Collaborative Efficiency}
\label{subsec:mas_planning}


Multi-agent systems (MAS) extend planning from individual deliberation to collaborative problem solving. 
Compared with single-agent planning, where efficiency mainly depends on how much reasoning, search, reflection, or decomposition one agent performs before acting, MAS introduces an additional coordination cost: agents must decide who should participate, what information should be exchanged, and when interaction should stop. 
Thus, the efficiency bottleneck shifts from controlling planning depth alone to jointly controlling depth and breadth.

This shift makes collaboration a double-edged sword. 
More agents or debate rounds can improve coverage and robustness, but they can also amplify message passing, lengthen contexts, and trigger redundant clarification or verification. 
Therefore, efficient MAS planning should not assume that larger teams or longer discussions always improve outcomes. 
Instead, it needs mechanisms that structure communication, compress exchanged information, route messages to relevant agents, and terminate interaction when the marginal benefit becomes low.

In this subsection, we review efficient MAS planning from three perspectives: \textbf{topological efficiency}, which reduces communication complexity by organizing agent interaction structures; \textbf{protocol optimization}, which controls the content and format of exchanged messages; and \textbf{distillation}, which transfers collaborative reasoning into cheaper single-agent or smaller-model execution.

\paragraph{Topological Efficiency and Sparsification.}
Topological efficiency optimizes the communication graph to mitigate quadratic message costs, often reducing message complexity from $\mathcal{O}(N^2)$ toward $\mathcal{O}(N)$ through structured topologies such as chains or DAGs.
\textit{Structured topologies} like Chain-of-Agents~\cite{zhang2024chain} and MacNet~\cite{qian2024scaling} restrict context growth to near-linear complexity, while GroupDebate~\cite{liu2024groupdebate} alternates between dense debate and sparse summaries.
\textit{Selective interaction} protocols further filter turns; MARS~\cite{wang2025mars} and S²-MAD~\cite{zeng2025s} reduce peer-to-peer noise by triggering debates only when viewpoints diverge.
More adaptive methods, such as AgentPrune~\cite{zhang2024cut}, AgentDropout~\cite{wang2025agentdropout}, and SafeSieve~\cite{zhang2025safesieve}, dynamically prune low-utility agents or edges during inference.
Beyond sparsification, InfoSeeker~\citep{lee2026infoseekerscalablehierarchicalparallel} uses a hierarchical Host--Manager--Worker structure for web information seeking, isolating context across layers while parallelizing independent evidence-gathering subtasks.
These methods suggest that efficient MAS design should jointly control communication topology, context scope, and execution parallelism.

\paragraph{Protocol and Context Optimization.}
Protocol optimization improves efficiency by compressing what is communicated, using concise representations such as pseudocode and prompt-driven constraints to reduce interaction context.
CodeAgents~\cite{yang2025codeagents} encodes reasoning in concise pseudocode, while Smurfs~\cite{chen2025smurfs} discards failed search branches to prevent context bloat. In parallel, prompt-level control accelerates convergence; Free-MAD~\cite{cui2025free} and ConsensAgent~\cite{pitre2025consensagent} engineer prompts to encourage critical reasoning, while supervisors like SMAS~\cite{lin2025stop} terminate redundant loops early.

\paragraph{Distilling Coordination into Planning.}
The most radical approach internalizes coordination by distilling collective intelligence into a single-agent model, bypassing runtime coordination costs.
Methods like MAGDI~\cite{chen2023magdi} and SMAGDi~\cite{aluru2025smagdi} distill complex interaction graphs or "Socratic" decomposition into a single student model. Similarly, D\&R~\cite{zhou2025debate} uses a teacher-student debate to generate preference trees for DPO. These approaches retain the quality benefits of diverse perspectives while reverting to the lower inference cost of a single agent.

\begin{summarybox}[]{Multi-Agent Collaborative Efficiency}
\begin{itemize}[leftmargin=0.5em]
\item \textbf{Pros}: Turns collaboration into controlled coordination, allowing agents to benefit from complementary expertise, parallel exploration, and mutual verification while limiting unnecessary communication.
\item \textbf{Cons}: Over-controlling collaboration may suppress useful disagreement, discard task-relevant context, or make the system too dependent on a fixed coordination pattern.
\item \textbf{Suitable situations}: Most useful when a task benefits from multiple perspectives or specialized agents, but unrestricted discussion would cause excessive token use, latency, or context growth.
\item \textbf{Trade-offs}: Multi-agent efficiency depends on the marginal value of communication: more agents and messages can improve coverage, but they also increase coordination cost, redundancy, and context pressure.
\end{itemize}
\end{summarybox}

\subsection{Discussion}

Efficient agent planning reframes reasoning from an unbounded generation process into a budget-aware control problem. In the \textbf{single-agent} regime, inference-time strategies decide how much computation to spend on the current problem, while learning-based evolution amortizes planning cost across future problems. In the \textbf{multi-agent} regime, efficiency depends on controlling the communication topology and protocol so that collaboration improves solution quality without turning coordination into the dominant cost. Across both settings, the unifying trend is the migration of computation from expensive \textit{online search} to cheaper \textit{offline learning}, reusable memory, or structured retrieval.

\paragraph{Planning as Meta-Control.}
A deeper way to view efficient planning is as meta-control: the planner must decide not only what action to take, but also how much reasoning, search, tool use, or communication should precede that action. This makes planning efficiency fundamentally different from simply shortening the generated reasoning trace. A short plan can be inefficient if it causes repeated failures, while a longer deliberation can be efficient if it prevents costly environment interactions or tool calls. Thus, the right unit of analysis is end-to-end cost, including tokens, wall-clock time, external actions, retries, and coordination overhead.

\paragraph{Depth, Breadth, and Amortization.}
The methods surveyed in this section expose three recurring efficiency axes. First, \textbf{depth control} decides how far a single agent should search, reflect, or decompose before acting. Second, \textbf{breadth control} decides how many agents, branches, or candidate plans should be explored in parallel. Third, \textbf{amortization} shifts useful planning knowledge into policies, skills, or memories so that future tasks require less online deliberation. These axes are complementary but also compete for budget: deeper single-agent search may reduce the need for multi-agent debate, while stronger memory or skill reuse may reduce both search depth and communication breadth.

\paragraph{When More Planning Becomes Inefficient.}
A key open question is how to identify the stopping point of planning. Additional search, reflection, or debate has diminishing returns: early deliberation may eliminate obvious errors, but later deliberation may mostly rephrase existing ideas or introduce new inconsistency. This is especially important in multi-agent systems, where additional agents can improve diversity but also create redundant messages and longer consensus processes. Efficient planners therefore need progress signals that estimate the marginal value of continued reasoning, rather than relying on fixed numbers of steps, agents, or debate rounds.

\paragraph{Coupling Planning with Memory and Tools.}
Planning sits between memory and tool use. Memory reduces planning cost by storing reusable subgoals, plans, failures, and skills; tools reduce planning burden by externalizing computation or verification. At the same time, planning determines when to retrieve memory, when to update memory, and when to call tools. This makes planning the controller of cross-component cost allocation: an efficient planner may spend more tokens upfront if doing so avoids repeated retrieval, unnecessary tool calls, or failed environment interactions. Conversely, better memory and tools can shorten planning by providing reusable evidence and executable operations. A planner should therefore be evaluated by whether it reduces total agent cost: fewer unnecessary tool calls, fewer repeated memory retrievals, fewer environment steps, and fewer multi-agent messages, while preserving task success.
\section{Benchmarks}
\label{sec:benchmark}
Although this survey focuses on efficiency, we adopt an effectiveness-first view that a method that is cheap but fails to solve tasks or substantially harms solution quality is not meaningfully efficient.
Accordingly, efficiency should not be evaluated in isolation. A practical evaluation protocol first establishes whether an agent or component remains effective, and then compares cost under a comparable effectiveness level, or compares effectiveness under a fixed cost budget. This trade-off can also be viewed through the Pareto frontier between effectiveness and cost.

A key challenge is that existing work rarely evaluates memory efficiency, tool efficiency, or planning efficiency with fully isolated benchmarks. Many methods instead report performance on holistic agent benchmarks, then add cost-related measurements. We therefore organize this section by evaluation purpose rather than by component. We first summarize benchmarks for measuring effectiveness, including holistic agent capability, downstream task performance, and component-specific ability, and then summarize efficiency measurements used in benchmarks and method papers. Table~\ref{tab:benchmark_taxonomy_example} gives an example of how these benchmarks and metrics can be tabulated.

\begin{table*}[t]
\centering
\footnotesize
\setlength{\tabcolsep}{5pt}
\renewcommand{\arraystretch}{1.18}
\begin{tabularx}{\textwidth}{>{\raggedright\arraybackslash}p{0.22\textwidth}|>{\raggedright\arraybackslash}X|>{\raggedright\arraybackslash}p{0.31\textwidth}}
\toprule
\textbf{Evaluation Scope} & \textbf{Representative Benchmarks} & \textbf{Typical Signals} \\
\midrule
\textbf{Holistic agent performance} & GAIA~\citep{mialon2023gaia}; SWE-Bench~\citep{jimenez2024swebench}; WebArena~\citep{zhou2023webarena}; WebShop~\citep{yao2022webshop}; $\tau$-Bench~\citep{yao2024tau}; $\tau^2$-Bench~\citep{barres2025tau2} & End-to-end task success, pass rate, answer correctness, and trajectory completion in interactive environments. \\
\addlinespace[2pt]
\textbf{Downstream task performance} & HotpotQA~\citep{yang2018hotpotqa}; Natural Questions~\citep{kwiatkowski2019natural}; SimpleQA~\citep{Wei2024MeasuringSF}; BrowseComp~\citep{Wei2025BrowseCompAS}; SealQA~\citep{Pham2025SealQART} & Final-output quality, including QA accuracy, multi-hop reasoning, factual correctness, and evidence synthesis. \\
\addlinespace[2pt]
\textbf{Memory-specific ability} & LoCoMo~\citep{maharana2024evaluating}; LongMemEval~\citep{wu2024longmemeval} & Long-term consistency, temporal reasoning, retention, and memory-grounded question answering. \\
\addlinespace[2pt]
\textbf{Tool-use-specific ability} & API-Bank~\citep{Li2023APIBankAC}; BFCL~\citep{patil2025bfcl}; MetaTool~\citep{huang2024metatoolbenchmarklargelanguage}; ToolBench~\citep{Qin2023ToolBench}; MGToolBench~\citep{wu2024toolplanner}; NesTools~\citep{han2025nestoolsdatasetevaluatingnested}; T-Eval~\citep{Chen2023TEvalET}; StableToolBench~\citep{qin2025stabletoolbenchtowardsstable}; MCP-RADAR~\citep{gao2025mcpradarmultidimensionalbenchmarkevaluating}; MCP-Bench~\citep{wang2025mcp} & Tool selection, parameter filling, schema following, execution correctness, protocol compliance, and redundant-call avoidance. \\
\addlinespace[2pt]
\textbf{Planning-specific ability} & PlanBench~\citep{valmeekam2023planbench}; Blocksworld-style planning~\citep{slaney2001blocks,jobs2025benchmark}; TPS-Bench~\citep{xu2025tps}; CostBench~\citep{liu2025costbench} & Plan validity, execution success, path quality, planning attempts, and recovery from invalid actions. \\
\addlinespace[2pt]
\textbf{Efficiency signals} & Evo-Memory~\citep{wei2025evo}; StoryBench~\citep{wan2025storybench}; MemBench~\citep{tan2025membench}; TPS-Bench~\citep{xu2025tps}; CostBench~\citep{liu2025costbench}; method-level evaluations & Tokens, latency, API cost, memory usage, tool calls, reasoning steps, and environment steps. \\
\bottomrule
\end{tabularx}
\caption{A taxonomy-style organization of benchmark coverage for evaluating efficient LLM agents. Holistic benchmarks measure full agent trajectories, downstream task benchmarks measure final-output quality, component-specific benchmarks diagnose memory, tool-use, or planning abilities, and efficiency signals are usually reported alongside effectiveness rather than as standalone scores.}
\label{tab:benchmark_taxonomy_example}
\end{table*}

\subsection{Effectiveness Benchmarks}
Effectiveness benchmarks answer the question of whether an agent, a downstream task setup, or a specific agent component can solve the target task at all. They are the necessary baseline for efficiency claims: reducing cost is only meaningful when the resulting system preserves task quality.

\paragraph{Holistic Agent Benchmarks.}
Many efficiency-oriented agent methods are evaluated on end-to-end benchmarks because memory, tool use, and planning interact tightly in realistic tasks. GAIA~\citep{mialon2023gaia} evaluates general assistant agents on questions requiring multi-step reasoning and tool use. Software, web, and shopping environments such as SWE-Bench~\citep{jimenez2024swebench}, WebArena~\citep{zhou2023webarena}, and WebShop~\citep{yao2022webshop} measure whether agents can complete realistic interactive tasks rather than only produce isolated answers. Tool-rich interactive benchmarks such as $\tau$-Bench~\citep{yao2024tau} and $\tau^2$-Bench~\citep{barres2025tau2} similarly evaluate end-to-end task completion in domains such as retail, airline, and telecom. These benchmarks are not tied to a single component, but they are often the most convincing testbed for showing that a proposed memory, tool-use, or planning mechanism improves overall agent performance.

\paragraph{Downstream Task Benchmarks.}
Some effectiveness benchmarks are neither holistic agent benchmarks nor component-specific diagnostic benchmarks. Instead, they evaluate final task performance and are often used to indirectly test whether memory, retrieval, or search-augmented tool use improves the final output. For memory evaluation, QA datasets such as HotpotQA~\citep{yang2018hotpotqa} and Natural Questions~\citep{kwiatkowski2019natural} can serve as indirect downstream tests, because better recall, retrieval, or context management can translate into more accurate answers. For search-augmented agents, SimpleQA~\citep{Wei2024MeasuringSF}, BrowseComp~\citep{Wei2025BrowseCompAS}, and SealQA~\citep{Pham2025SealQART} evaluate difficult fact-seeking questions where agents may need to retrieve and synthesize external evidence. These benchmarks are useful for measuring task-level effectiveness, but they should be distinguished from holistic agent benchmarks that evaluate full interactive trajectories.

\paragraph{Memory Effectiveness.}
Agent memory effectiveness can also be evaluated with memory-targeted tasks~\citep{zhang2025survey}. Unlike downstream QA benchmarks, these benchmarks more directly isolate whether an agent can store, maintain, and use information over long contexts or conversations. LoCoMo~\citep{maharana2024evaluating} and LongMemEval~\citep{wu2024longmemeval} are representative examples that evaluate long-term retention, temporal reasoning, and memory-grounded question answering.

\paragraph{Tool-Use Effectiveness.}
Tool-use benchmarks evaluate whether an agent can decide when to call tools, choose appropriate tools, fill parameters correctly, and execute multi-step tool workflows. Seal-Tools~\citep{wu2024sealtoolsselfinstructtoollearning} and UltraTool~\citep{huang2024planningcreationusagebenchmarking} study tool construction and use-case generation, which affect the quality of downstream tool-use behavior. MetaTool~\citep{huang2024metatoolbenchmarklargelanguage} focuses on the decision of whether to use a tool and which tool to select. API-Bank~\citep{Li2023APIBankAC} and the Berkeley Function-Calling Leaderboard (BFCL)~\citep{patil2025bfcl} evaluate function calling, parameter filling, and schema adherence in realistic tool scenarios. ToolBench~\citep{Qin2023ToolBench} scales evaluation to a large API collection, while MGToolBench~\citep{wu2024toolplanner} curates ToolBench with multiple granularities to better reflect real user queries. NesTools~\citep{han2025nestoolsdatasetevaluatingnested} targets nested and compositional tool calls. T-Eval~\citep{Chen2023TEvalET} decomposes tool utilization into fine-grained capabilities such as planning, reasoning, and retrieval, making failures easier to localize than with a single end-to-end score. StableToolBench~\citep{qin2025stabletoolbenchtowardsstable} improves reproducibility by replacing unstable online APIs with a virtual API server and simulation-based evaluation.

As tool ecosystems become standardized, protocol-level benchmarks have also emerged. MCP-RADAR~\citep{gao2025mcpradarmultidimensionalbenchmarkevaluating} evaluates agents under the Model Context Protocol (MCP), including tool selection, execution, and efficiency-related dimensions. MCP-Bench~\citep{wang2025mcp} evaluates MCP agents with an LLM-as-a-Judge rubric, including whether the agent exploits parallelism and avoids redundant calls.

\paragraph{Planning Effectiveness.}
Planning benchmarks evaluate whether agents can construct valid action sequences and recover from long-horizon dependencies. PlanBench~\citep{valmeekam2023planbench} studies planning in more controlled LLM settings, while agentic environments such as SWE-Bench, WebArena, WebShop, $\tau$-Bench, and $\tau^2$-Bench implicitly test planning through interactive task completion. Blocksworld-style domains~\citep{slaney2001blocks} remain useful for controlled planning analysis, and recent agent-planning benchmarks such as the structured benchmark of \citet{jobs2025benchmark}, TPS-Bench~\citep{xu2025tps}, and CostBench~\citep{liu2025costbench} evaluate planning and execution under explicit task constraints. These benchmarks are especially relevant for efficiency because poor plans often increase the number of tool calls, environment interactions, or retries even when the final answer is correct.

\subsection{Efficiency Measurements}
Efficiency measurements answer the question of how much cost is required to reach a given level of effectiveness. Unlike effectiveness, efficiency is often not tied to a single benchmark family. Instead, papers report cost-related metrics alongside downstream performance, and these metrics can be grouped into token and monetary cost, time cost, resource cost, and interaction cost.

\paragraph{Benchmark-Level Efficiency Signals.}
Some benchmarks explicitly include efficiency-related measurements. Evo-Memory~\citep{wei2025evo} introduces step efficiency, measuring how many environment steps are required to reach a goal. StoryBench~\citep{wan2025storybench} reports runtime cost and token consumption for long-horizon memory-augmented tasks. MemBench~\citep{tan2025membench} reports read time and write time in seconds per memory operation, directly reflecting the overhead of memory maintenance. For planning and tool use, the structured Blocksworld-based benchmark proposed by \citet{jobs2025benchmark} reports execution time, planning attempts, token consumption, and monetary cost. TPS-Bench~\citep{xu2025tps} evaluates planning and tooling efficiency with token usage, end-to-end time, and tool-call turns, and further proposes cost-of-pass to connect monetary cost with completion rate. CostBench~\citep{liu2025costbench} models tools with explicit costs and measures Cost Gap and path deviation, while also counting invalid tool calls as a source of tooling inefficiency. MCP-RADAR and MCP-Bench also include efficiency-oriented dimensions, such as computational resource efficiency, execution speed, redundant tool-call avoidance, and parallelism.

\paragraph{Token and Monetary Cost.}
Token consumption is the most common efficiency signal because it approximates both context-processing overhead and API cost. Memory methods frequently report token usage or token savings, including RECOMP~\citep{xu2023recomp}, GraphReader~\citep{li2024graphreader}, HiAgent~\citep{hu2025hiagent}, SeCom~\citep{pan2025secom}, Mem0~\citep{chhikara2025mem0}, A-MEM~\citep{xu2025mem}, MemoryOS~\citep{kang2025memory}, and MEM1~\citep{zhou2025mem1learningsynergizememory}. Some works translate token usage into dollar cost, such as Agentic Plan Caching~\citep{zhang2025costefficientservingllmagents} and ACE~\citep{zhang2025agentic}. Planning methods also commonly report token consumption, including SwiftSage~\citep{lin2023swiftsagegenerativeagentfast}, Language Agent Tree Search~\citep{zhou2024languageagenttreesearch}, ReWOO~\citep{xu2023rewoo}, and ReSo~\citep{zhou2025resorewarddrivenselforganizingllmbased}. Cost-of-pass style metrics~\citep{erol2025cost} further combine cost with success probability, and have been adopted or extended in TPS-Bench and subsequent work~\citep{wang2025efficient,wu2025gap}. Other studies compare performance under the same budget, such as budget-aware tool use, QLASS, and trial-and-error style approaches~\citep{liu2025budgetawaretooluseenableseffective,lin2025qlass,song2024trialanderroreto}.

\paragraph{Time and Runtime Overhead.}
Time-based metrics capture latency and runtime overhead, which are crucial when agents perform repeated retrieval, tool calls, or planning loops. HiAgent reports overall runtime, while SeCom, Mem0, and MemOS~\citep{li2025memos} measure end-to-end latency that combines search and reasoning but excludes construction time. MEM1 reports inference time. More fine-grained retrieval overhead is measured by A-MEM, H-MEM~\citep{sun2025hierarchical}, and Agent KB~\citep{tang2025agent}, which report retrieval time or search latency. MemoRAG~\citep{qian2025memorag} further distinguishes index latency from retrieval latency. In planning, ToolChain~\citep{zhuang2023toolchainefficientactionspace} reports runtime, and benchmark-level studies such as TPS-Bench and CostBench connect runtime with execution success.

\paragraph{Resource Cost.}
Resource-based metrics quantify hardware consumption and system overhead. A-MEM and MemoRAG report GPU memory usage, and MemoRAG additionally analyzes GPU memory consumption under different context lengths. MCP-RADAR also includes computational resource efficiency in its protocol-level evaluation. These metrics are especially important for latent-memory and retrieval-heavy systems, where improvements in token usage may come at the expense of larger caches, indexes, or memory stores.

\paragraph{Interaction and Search Cost.}
Interaction-based metrics measure how intensively an agent interacts with the LLM, tools, or environment. MemoryOS reports the average number of LLM calls per response, while ReasoningBank~\citep{ouyang2025reasoningbank} tracks the number of reasoning steps. Planning methods often report search-depth or search-breadth indicators: SwiftSage counts time steps, Reflexion~\citep{shinn2023reflexionlanguageagentsverbal} counts trials, Language Agent Tree Search reports the number of explored nodes or states, and CATS~\citep{zhang2025costaugmented} reports the average number of iterations needed to find a valid solution. These metrics are closely tied to practical efficiency because extra reasoning steps, tool calls, retries, or environment actions directly increase latency and cost.

Overall, the benchmark landscape suggests that effectiveness and efficiency should be reported jointly. Holistic agent benchmarks reveal whether a method improves real task completion, downstream task benchmarks measure final-output quality, component-specific benchmarks diagnose where the improvement comes from, and efficiency metrics explain whether the same capability is achieved with fewer tokens, less time, lower hardware cost, or fewer interactions.

\section{Challenges and Future Directions}
\label{sec:challenge}
\paragraph{Toward Unified Efficiency Evaluation for Agent Systems.}
Unified efficiency evaluation is still missing for agents, as existing work reports inconsistent metrics and often collapses efficiency to token or API cost, with unclear boundaries for runtime, latency, memory, tool-use, and planning overhead.
A key next step is to establish a unified efficiency evaluation framework for agents. Current studies use inconsistent dimensions and terminology, often reduced to token usage or API cost, while other metrics such as runtime, latency, and step efficiency are reported without clear pipeline boundaries. Standardized accounting for memory, tool use, and planning overhead remains limited. Developing common stage definitions, metric granularity, and reporting protocols would make results comparable and clarify cost–performance trade-offs across agent designs.

\paragraph{Agentic Latent Reasoning.} Recent months have seen growing interest in latent-space reasoning for LLM\citep{wang2025system,li2025seekdarkreasoningtesttime,xu2025softcotsoftchainofthoughtefficient}
, where intermediate computations are carried out in continuous hidden representations rather than being fully externalized as natural-language tokens. Compared with token-level “decode-and-read” reasoning, latent reasoning can reduce token overhead and may preserve richer, high-dimensional information during multi-step computation. However, existing work has largely focused on standalone LLM settings, while agentic latent reasoning remains relatively underexplored. This gap is important because agentic scenarios introduce additional requirements, such as tool use, long-horizon planning, memory management, and action verification, that differ from pure text-only reasoning and may demand new training objectives, interfaces, and evaluation protocols. Investigating latent reasoning mechanisms tailored for agents could therefore be a promising future direction.

\paragraph{Deployment-Aware Agentic Design.} 
Inspired by MemAgent~\citep{yu2025memagent} and Chain-of-Agents~\citep{zhang2024chain}, which address long-context reasoning by chunking context and processing it sequentially, we argue that agentic systems should be more deployment-aware. In practice, multi-agent designs can be realized either as true multi-model deployments or as single-model role-play pipelines, and these implementations differ substantially in orchestration overhead, latency, and reliability. Future work should compare these alternatives under matched resource budgets and report end-to-end cost–benefit metrics, clarifying whether the performance gains from adding more agents justify the additional complexity.



\paragraph{Efficiency Challenges and Directions for MLLM-based Agents.}
There has been a rapid emergence of MLLM-based agent methods, including agents equipped with multimodal memory~\citep{li2024optimus,wang2024videoagent,sarch2024vlm}, approaches that explicitly enhance planning and decision-making for MLLM-based agents~\citep{liu2025infiguiagentmultimodalgeneralistgui}, and multi-agent systems built upon LLM and MLLM backbones~\citep{yang2025llm,wang2025mirix}, among others. However, efficiency in MLLM-based agents is relatively under-explored, which is also emphasized in~\citep{yao2025survey}. In realistic deployments, efficiency is crucial due to the need for rapid responses under strict latency and compute budgets.
In this regard, we observe that several efficiency techniques in LLM-based agents may inspire MLLM-based agents. For example, SwiftSage~\citep{lin2023swiftsagegenerativeagentfast} adopts fast–slow mode switching to allocate computation adaptively, and FAST-GRPO~\citep{xiao2025fast} explores a similar fast–slow thinking mechanism for MLLM agents.
Nevertheless, transferring text-centric efficiency strategies to multimodal agents remains challenging. Compared with language-only settings, MLLM-based agents often operate in different action spaces and task structures, such as GUI-based or embodied interactions, while multimodal perception and grounding can introduce additional latency and compound errors over long-horizon interactions \cite{he2025diffthinker}. Notably, long-horizon multimodal tasks require maintaining a visual history. The cumulative computational burden of re-encoding visual context for every step creates a trade-off between memory retention and inference speed that is far more severe than in LLM-based agents.
As a future direction, we advocate efficiency-aware agent design and evaluation for MLLM-based agents by jointly considering performance and cost, including latency, interaction steps, and tool-call overhead.
\section{Conclusion}
\label{sec:conclusion}
In conclusion, this survey summarizes the evolution from LLMs to
LLM-based agents, highlighting the shift toward increasingly complex settings that
motivates our discussion. We review three core components, \textit{memory},
\textit{tool use}, and \textit{planning}, with an emphasis on efficiency, and find
that many seemingly different methods converge on shared high-level ideas. We also
summarize efficiency-oriented benchmarks and commonly used metrics across both benchmark
and methodological studies. Finally, we outline key challenges and future directions. Overall, our survey consolidates the design space and
evaluation practices for agent efficiency, while underscoring the need for more
standardized and transparent reporting to enable fair comparison and reproducibility.
We hope this survey offers useful guidance for designing and evaluating efficient agents, and helps encourage further progress in this direction.



\bibliographystyle{plainnat}
\bibliography{ref}

\end{document}